\RequirePackage[svgnames]{xcolor}

\documentclass[11pt,a4paper]{mystyle}

\usepackage[all]{hypcap}
\usepackage[svgnames]{xcolor}
\usepackage[comma,authoryear,compress]{natbib}
\renewcommand{\cite}{\citep}

\usepackage{algorithm}
\usepackage{algorithmicx}
\usepackage{algpseudocode}
\usepackage{microtype}
\usepackage{booktabs}
\usepackage{float}
\usepackage{bigstrut}
\usepackage{multirow}

\usepackage{amsmath}
\usepackage{amssymb}
\usepackage{mathtools}
\usepackage{amsthm}
\usepackage{mathrsfs}
\usepackage{dsfont}
\usepackage{enumitem}
\usepackage{graphicx}
\usepackage{cleveref}
\usepackage{bxcoloremoji}

\usepackage{float}

\usepackage[utf8]{inputenc} %
\usepackage[T1]{fontenc}    %
\usepackage{hyperref}       %
\usepackage{url}            %
\usepackage{booktabs}       %
\usepackage{amsfonts}       %
\usepackage{nicefrac}       %
\usepackage{microtype}      %
\usepackage{amssymb}
\usepackage{fdsymbol}
\usepackage{wrapfig}
\usepackage{lipsum}
\usepackage{enumitem}
\usepackage{stackengine}
\usepackage[font=small,labelfont=bf]{caption}
\usepackage{color}
\usepackage{adjustbox}

\usepackage{rotating}
\usepackage{makecell}

\title{LLM-Generated Fake News Induces Truth Decay in News Ecosystem: A Case Study on Neural News Recommendation}

\runningtitle{LLM-Generated Fake News Induces Truth Decay in News Ecosystem: A Case Study on Neural News Recommendation}

\author[1,2]{\href{https://beanandrew.github.io/}{\textcolor{black}{Beizhe Hu}}}
\author[1]{\href{https://sheng-qiang.github.io/}{\textcolor{black}{Qiang Sheng}}}
\author[1,2]{\href{https://scholar.google.com/citations?user=fSBdNg0AAAAJ}{\textcolor{black}{Juan Cao}}}
\author[1,2]{\href{https://liesy.github.io/}{\textcolor{black}{Yang Li}}}
\author[1]{\href{https://scholar.google.com/citations?user=hGZwK0cAAAAJ}{\textcolor{black}{Danding Wang}}}

\affil[1]{Media Synthesis and Forensics Lab, Institute of Computing Technology, Chinese Academy of Sciences}
\affil[2]{University of Chinese Academy of Sciences}

\correspondingauthor{Qiang Sheng}
\emails{{\{\href{mailto:hubeizhe21s@ict.ac.cn}{hubeizhe21s},\href{mailto:shengqiang18z@ict.ac.cn}{shengqiang18z},\href{mailto:caojuan@ict.ac.cn}{caojuan},\href{mailto:liyang23s@ict.ac.cn}{liyang23s},\href{mailto:wangdanding@ict.ac.cn}{wangdanding}\}@ict.ac.cn}}

\begin{document}

\begin{abstract}
Online fake news moderation now faces a new challenge brought by the malicious use of large language models (LLMs) in fake news production.
Though existing works have shown LLM-generated fake news is hard to detect from an individual aspect, it remains underexplored how its large-scale release will impact the news ecosystem.
In this study, we develop a simulation pipeline and a dataset with \textasciitilde 56k generated news of diverse types to investigate the effects of LLM-generated fake news within neural news recommendation systems.
Our findings expose a truth decay phenomenon, where real news is gradually losing its advantageous position in news ranking against fake news as LLM-generated news is involved in news recommendation.
We further provide an explanation about why truth decay occurs from a familiarity perspective and show the positive correlation between perplexity and news ranking.
Finally, we discuss the threats of LLM-generated fake news and provide possible countermeasures. We urge stakeholders to address this emerging challenge to preserve the integrity of news ecosystems.
\vspace{5mm}

\coloremojicode{1F4C5} \textbf{Date}: April 28, 2025

\coloremojicode{1F3E0} \textbf{Project}: \href{https://beanandrew.github.io/projects/TruthDecay}{https://beanandrew.github.io/projects/TruthDecay}

\coloremojicode{1F4AC} \textbf{Venue}: ACM SIGIR 2025 (\href{https://doi.org/10.1145/3726302.3730027}{https://doi.org/10.1145/3726302.3730027})

\end{abstract}

\maketitle
\vspace{3mm}

\begin{figure}[ht]
	\centering
	\includegraphics[width=0.63\textwidth]{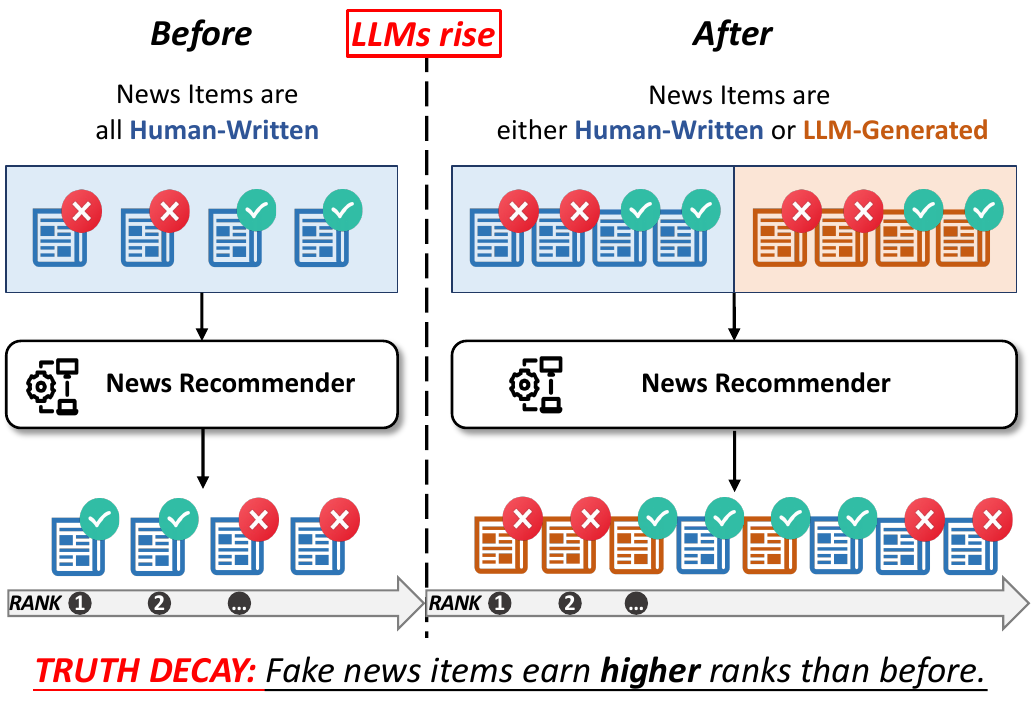}
	\caption{Illustration of Truth Decay phenomenon, where real news gradually \textit{loses} its advantageous position in rankings against fake news as LLM-generated news penetrates.}
	\label{fig:main_fig}
\end{figure}

\section{Introduction}

The proliferation of online fake news has been a long-standing problem across the world, posing significant threats across various critical domains such as politics~\cite{science19} and public health~\cite{caceres2022impact}.
Fake news moderation has been a necessity for online information systems like social media~\cite{giansiracusa2021social}, search engines~\cite{google-fact-check}, and recommendation systems~\cite{fernandez2020recommender}.
To combat the fake news spread, stakeholders have made much effort in developing automatic detection techniques~\cite{hu2022deep,bu2023combating,m3fend}, monitoring and mitigating fake news spread~\cite{saxena2022fake}, providing reporting and debunking systems~\cite{refutation}, which effectively reduce fake news threats to the real world.
Unfortunately, the breakthrough in text generation with large language models (LLMs) has largely complicated the situation~\cite{zhao2024survey,chen2024combating}.
Because of their excellent text generation capability, malicious actors have equipped themselves with LLMs to cheaply and efficiently produce fake news~\cite{jiang2024disinformation}. 
Recent coverage discloses that over 1k unreliable AI-generated news websites were identified~\cite{newsguard} globally, and a news faking gang using LLMs has produced 28k fake news items, surprisingly resulting in 2.7B views~\cite{wenzhou}.
As LLM-generated news \textit{rushes} into online news channels, the news ecosystem, like recommendation systems, faces an almost unavoidable news data distribution shift and credibility crisis.
A crucial but still underexplored question arises: \textbf{How does LLM-generated fake news impact the neural news recommendation system?}

To investigate the impact on news recommendation systems, we first tackle the unavailability of existing data and perform \textbf{Data \& Environment Preparation}.
By repurposing a fake news detection dataset GossipCop, we construct a dataset including news items, veracity labels, and user-news interaction records.
To systematically analyze LLM-generated news, we first propose a news taxonomy for levels of LLMs' news generation modes, given the degree of LLM involvement in generation.
Then we prompt LLMs to generate news of different types with the assigned level as guidance and human news as reference.
With this prepared data, we explore \textbf{How Generated News Affects the Neural News Recommendation System}. We initially test the system recommendation preference between human-written real and fake news, finding that overall, real news ranks higher than fake news. Subsequently, we examine the effects of LLM-generated news on the news recommendation system by simulating its intrusion into different components, including the news candidate list, user-news interaction history, and training set of news recommendation models.
As illustrated in \figurename~\ref{fig:main_fig}, we find that \textbf{LLM-generated news's involvement leads to the ``Truth Decay'' phenomenon, where real news is gradually losing its advantageous position against fake news and fake news (especially the LLM-generated) could be even more preferable and earns higher ranks than real news.}
Finally, we provide an analysis about \textbf{Why Truth Decay Occurs} from a familiarity perspective and reveal the perplexity's high correlation with the changes of ranking advantages. We further discuss the potential implications of the truth decay phenomenon and possible countermeasures.
Our main contributions include:
\begin{itemize}
[nosep,leftmargin=1em,labelwidth=*,align=left]
    \item \textbf{New Perspective:} To the best of our knowledge, we are the first that investigate the impact of LLM-generated fake news on neural news recommendation systems, which is important for credibility analysis of information systems in the LLM era.
     \item \textbf{New Findings:} We observe a \textbf{Truth Decay} phenomenon in neural news recommendation systems, which reflects that the involvement of LLM-generated news leads to relative advantages of real news over fake news being diminished and the overall deterioration, sounding a new alarm about such potential harms.
    \item \textbf{New Taxonomy \& Data:} We propose a new taxonomy for levels of LLM-based news generation modes given the degree of LLM involvement in generation, which facilitates systematic description and analysis of possible scenarios of LLM-based news generation. A human-LLM-mixed news recommendation dataset containing \textasciitilde75k news items that follow this taxonomy is constructed.  
\end{itemize}

\section{Preliminaries}

\subsection{Personalized News Recommendation}

\begin{figure}[h]
	\centering
	\includegraphics[width=0.8\textwidth]{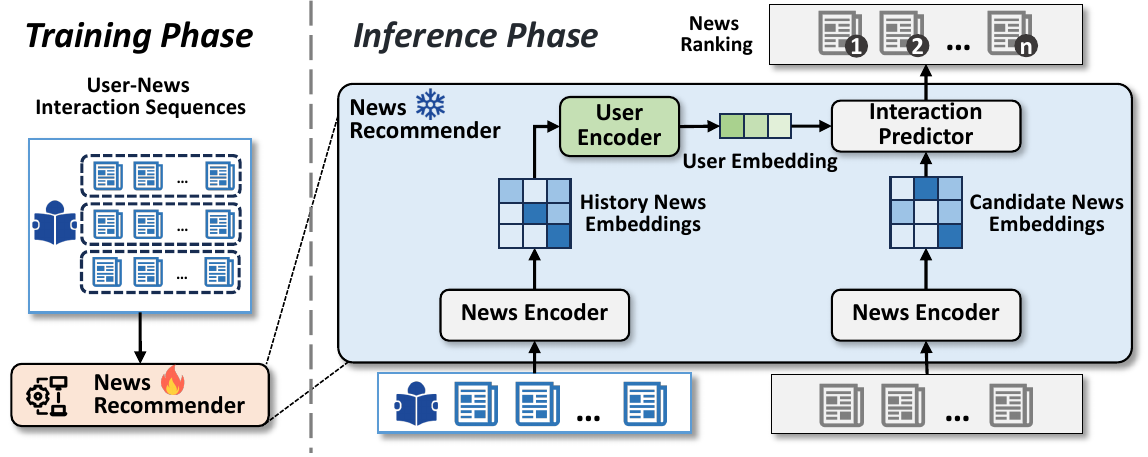}
	\caption{Illustration of a typical news recommendation model in the (Left) training and (Right) inference phases.}
	\label{fig:newrec_fig}
\end{figure}

Personalized news recommendation aims at actively displaying news items that might be of a target user's interest according to its news interaction history, mostly applied in news distribution channels like Google News\footnote{\url{https://news.google.com/foryou}}, MSN\footnote{\url{https://microsoftnews.msn.com/}}, Toutiao\footnote{\url{https://en.wikipedia.org/wiki/Toutiao}}, etc.
Given a user's news interaction history (\textit{e.g.,} viewing or reposting records), the news recommendation model predicts the next news item the user will interact with.
Like many other recommendation systems, news recommendation could work with collaborative filtering~\citep{cf1,cf2}, which predicts the next item according to other similar users' news interaction history. However, it faces a serious challenge of cold-start situations for news recommendation because news pools are intrinsically updated fast to ensure timeliness, causing interaction records on outdated news items to be hardly useful.
Therefore, content-based methods, most of which use neural networks for content representation~\citep{NRMS, LSTUR, PLM4REC, MINER}, 
are often considered effective for addressing the unique challenges of news recommendation systems.
\figurename~\ref{fig:newrec_fig} illustrates how a typical content-based news recommendation model is trained and then serves for ranking candidate news items.
During the training phase, the model
learns to predict whether an item in the candidate news will be interacted with based on the recently interacted news list. 
For inference, the news recommendation model is frozen, and candidate news items will be scored and ranked by referring to a user's interaction history. The ranked list will finally determine the display order of candidate news items to the target user.

Specifically, given a user's news interaction history and candidate news list, the model first employs news encoders to obtain history and candidate news embeddings. The history news embeddings are then transformed through a user encoder into a user embedding. The user embedding and candidate news embeddings are fed into an interaction predictor, which assigns a score to each news item. These scores are used to rank the candidate news.
Among the existing news recommendation works, LSTUR~\cite{LSTUR} and NRMS~\cite{NRMS} are one of the pioneering and well-known ones that adopt this design. 
Built upon this framework, some subsequent works improve recommendation performance by introducing topic information~\cite{TADI}, user interests~\cite{MINER}, or adding auxiliary pre-training tasks~\cite{PUNR}.

\subsection{Related Work}
\noindent\textbf{LLM-Generated Text Propagation in Information Systems.}
With the increasing prevalence of LLM-generated text, researchers attempt to estimate its propagation and influence in information systems of various domains. In an analysis of the scientific peer review system, \citet{monitoring} claimed that 6.9\%-16.9\% of review comments in post-ChatGPT AI conferences might be LLM-generated. \citet{dai2024cocktail,dai2024neural} and \citet{zhou2024source} discovered that neural information systems have an intrinsic preference for LLM-generated text over human-written text. \citet{spiral} simulated the impact of LLM-generated text on retrieval-augmented generation systems, identifying a phenomenon where LLM-generated text gradually dominates the system and thus marginalizes human-written content.
Differently, our work aims to reveal the impact of LLM-generated text on information credibility issues in news ecosystems, with a special focus on neural news recommendation.

\noindent\textbf{Fake News Propagation and Mitigation in Information Systems.} Besides developing detection techniques, researchers also pay attention to the spread of human-written fake news in online information systems.
\citet{fernandez2020recommender} described the critical issues and challenges in measuring the impact of recommendation algorithms on fake news dissemination. \citet{fernandez2021analysing} examined how popular recommendation algorithms contribute to the spread of fake news, while \citet{TWEB23Contribution} investigated the susceptibility of different algorithms to fake news and simulated its propagation across social networks.
To mitigate fake news spread, some works integrate credibility assessment capability into recommendation systems.
\citet{rec4mit} proposed a veracity-aware framework for news recommendation, which not only captures user preferences but also predicts the news veracity, thereby maintaining recommendation quality while avoiding promoting fake news. 
\citet{HDInt} proposed a framework that incorporates a veracity disentangler and classifier to identify and filter fake news, ensuring recommendations are both accurate and truthful.
Moreover, \citet{victor} developed a module that suggests verified, related news to users who have previously encountered false information, therefore mitigating the cognitive impact of fake news.
In this work, we expand the investigation scope from human-written fake news to LLM-generated fake news via a propagation simulation to reflect the new, possible threat in the near future. 

\noindent\textbf{Effects of LLM-Generated Fake News.}
Large language models like GPT-3~\cite{gpt3} and its successors have been empirically shown to be capable of generating seemingly high-quality news articles with careful guidance~\cite{zellers2019defending,huang2023faking,wang2024mega,aipress}.
However, LLMs' internal hallucination plus external induction makes their generated news mostly unreliable~\cite{chen2024combating,goldstein2024how}.
Moreover, recent works have shown that the generated fake news is hard to detect for both humans~\cite{spitale2023ai,chen2024can,wang2024reopening} and models trained on previous human news data~\cite{su2023fake,chen2024can} and could be even more deceptive with adversarial strategies added~\cite{wu2024fake,park2024adversarial,sun2024exploring}.
Different from previous works, we do not observe the individual impacts of LLM-generated fake news items but put them into a neural news recommendation system to expose how LLM-generated fake news competes with others for higher ranks and visibility, which ultimately reflects its impact on the news ecosystem.

\noindent\textbf{}

\section{Data \& Environment Preparation}

\begin{figure}[h]
	\centering
	\includegraphics[width=0.6\textwidth]{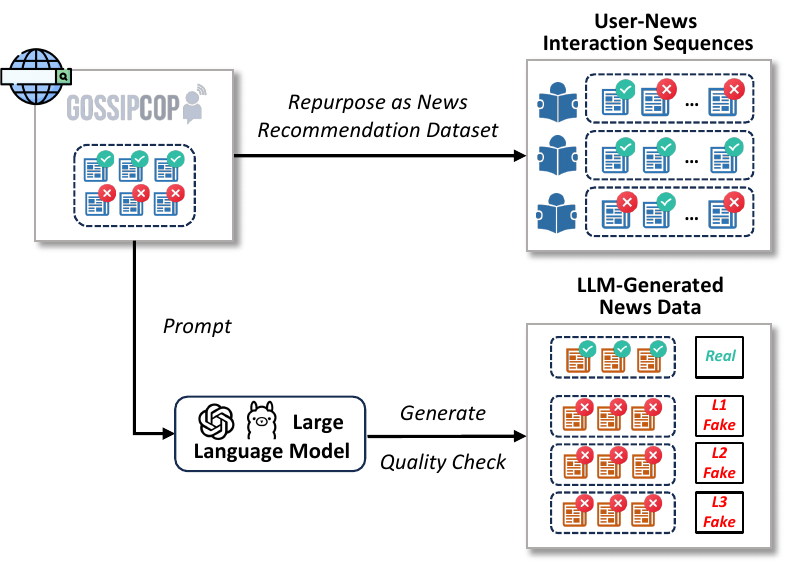}
	\caption{Illustration of dataset construction process, which consists of (Top) repurposing a fake news detection dataset for news recommendation and (Bottom) prompting LLMs to generate news data and performing quality check.}
	\label{fig:data_fig}
\end{figure}

Our investigation requires a dataset that includes both real and fake news items written by humans or generated by LLMs with user-news interaction behavior records. Given that no publicly available dataset satisfies the mentioned requirement, as shown in \figurename~\ref{fig:data_fig}, we construct a new dataset by 1) repurposing an existing human news dataset originally for fake news detection, 2) prompting two LLMs to generate news items according to the human news data and 3) performing a quality check. The new dataset will support the simulation of LLM-generated news involvement in news recommendation in Section~\ref{sec:effects}.

\subsection{Construction of News Recommendation Dataset with Veracity Labels}

In news recommendation, public datasets such as MIND~\citep{mind} and Adressa~\citep{adressa} lack veracity labels, while conventional fake news detection datasets do not directly provide user-news interaction sequences. Following \citet{rec4mit}, \citet{TWEB23Contribution}, and \citet{HDInt}, we construct a news recommendation dataset with the veracity of each news item labeled by repurposing a fake news detection dataset with social context information.
Specifically, our construction is based on the widely-used GossipCop dataset~\citep{fakenewsnet}, which compiles news articles fact-checked by the GossipCop website and their spread records on Twitter (now X).\footnote{Besides GossipCop, we have considered Politifact~\citep{fakenewsnet}, ReCOVery~\citep{Recovery}, and FakeHealth~\citep{FakeHealth} datasets but finally do not use them because of the insufficiency of individual news items, sparsity of user-item interactions, and difficulty in data access.} Therefore, we could obtain news veracity labels by transforming the fact-checking conclusion into a veracity class space (real/fake) and the news-centric spread records into user-centric interaction records with news items.
Here we consider users' tweeting, retweeting, and commenting actions as valid interactions, and construct a user-news interaction sequence for each user by sorting these interactions chronologically according to the given timestamps.

We further divide each user's interaction sequence into multiple sequence instances. For the interaction sequences of the user \(i\), denoted as \(S_i = [n_1, n_2, \ldots, n_T]\), where \(n_t\) represents the \(t\)-th news item in the user's sequence, we define a minimum interaction history \(H_{\text{min}}\). For each \(t\) satisfying \(H_{\text{min}}+1 \leq t \leq T\), \(n_t\) is taken as the target interacted news, with the sequence \([n_1, n_2, \ldots, n_{t-1}]\) serving as the user's interaction history.
Let \(K\) be the number of news items in a candidate list. For each target news item \(n_t\), we prepare additional \(K-1\) news items that the user has not interacted with previously. If \(n_t\) is a real news item, the candidate set includes \(K/2-1\) real news items and \(K/2\) fake news items; conversely, if \(n_t\) is fake, the set includes \(K/2-1\) fake news items and \(K/2\) real news items. Consequently, we derive a news recommendation dataset consisting of these user-news interaction sequence instances.

Specifically, we set the minimum interaction history \( H_{min}\) to 4, the candidate list length \( K\) to 20, and capped the user interaction history at 10. 
Following~\cite{TWEB23Contribution,mind}, we filtered out users with fewer than five interactions to ensure data quality.
For users with more than 10 user-interaction sequence instances, we randomly selected 10 consecutive instances.
We randomly selected 5,000 users' interaction sequences considering the simulation cost.
To avoid the interactions in the training set overlapping in time with those in the validation and test sets, we applied a temporal split strategy, using the most recent 20\% of interactions for the validation and test sets, while the remaining earlier 80\% serve as the training set. Table~\ref{tab:dataset_info} presents the statistics of the repurposed dataset.

\begin{table}[h]
  \centering
  \caption{Statistics of the news recommendation dataset.}
        \begin{tabular}{ccccccc}
        \toprule
        \multicolumn{2}{c}{\textbf{Behavior Record}} & \multicolumn{2}{c}{\textbf{News Item}} & \multicolumn{3}{c}{\textbf{Subset}} \\
        \cmidrule(lr){1-2} \cmidrule(lr){3-4} \cmidrule(lr){5-7}
        \# User & \# Interaction & \# Real & \# Fake & \# Train & \# Validation & \# Test \\
        \cmidrule(lr){1-2} \cmidrule(lr){3-4} \cmidrule(lr){5-7}
        5,000  & 46,910 & 14,248 & 4,592  & 18,761 & 2,346  & 2,345 \\
        \bottomrule
        \end{tabular}
  \label{tab:dataset_info}
\end{table}

\subsection{LLM-Generated News Dataset Construction}

\noindent\textbf{Taxonomy of LLM-Based News Generation Modes.}
By referring to real-world cases and existing works~\cite{chen2024can,wu2024fake}, we compile a taxonomy of LLM-based news generation modes based on the degree of LLM involvement in generation. Specifically, we focus on the following four modes.\footnote{This mimics the SAE's taxonomy of driving automation systems~\cite{sae}. We omit L4/L5 here (Appendix~\ref{app:l4l5}) as their practice is still limited by LLMs' current capabilities.}
\begin{itemize}
[nosep,leftmargin=1em,labelwidth=*,align=left]
     \item \textbf{L0: No Generator Automation (Human-Written).} The news piece is purely written by a human writer without any involvement of LLM generators.
    \item \textbf{L1: Human Assistance.} This is the lowest level of generation automation, where the generator provides simple enhancements to human-written news while preserving the original content and style. Specifically, given a piece of human-written news, the LLM is prompted to paraphrase it.
    
    \item \textbf{L2: Partial Generator Automation.} At this level, the generator has more freedom, modifying the style of human-written news while maintaining the core content. This results in news with a significantly altered style and some content changes. Specifically, given a piece of human-written news, the LLM is prompted to rewrite it to make it more convincing.
    
    \item \textbf{L3: Conditional Generator Automation.} This level allows the generator even greater autonomy to create content based on provided news materials, resulting in significant changes in both content and style while retaining the news theme. Specifically, given the title and main content of a human-written news piece as material, the LLM is prompted to create a convincing news item based on these inputs.
\end{itemize}

\noindent\textbf{Generation Details.}
To construct the LLM-generated news dataset, we used the L1, L2 and L3 modes mentioned before. Specifically, we employed all three generation modes to create LLM-generated fake news, and utilized generation mode L1 to generate LLM-generated real news, as modes L2 and L3 could alter text content, making it difficult to maintain the veracity of the news. Considering the potentially huge cost of further verification, we assume that modes L2 and L3 are unlikely to be practiced to generate real news in the status quo of real-world scenarios and thus do not consider generating real news at L2 and L3. Hereafter, we may call generation modes as \textit{faking levels} if the modes are applied to fake news generation.

Specifically, we employed gpt-4o-mini (\textit{gpt-4o-mini-2024-07-18})  and Llama-3.1 (\textit{Llama-3.1-8B-Instruct}), which are representative of open-source and proprietary LLMs, respectively, to construct the dataset. In the generation process, we set the temperature to 0.7 and the maximum generation length to 2,048 tokens. In the post-processing step, we identified and removed invalid and news-irrelevant texts and unnecessary symbols from the LLM responses. For example, some LLM outputs usually start with the reply that expresses the acceptance of a user's instruction (\textit{e.g.,} ``Sure''). \tablename~\ref{tab:generation_types_fig} shows the generation prompts and cases, and \tablename~\ref{tab:llm_data} presents the statistics of LLM-generated news data.

\begin{table}[h]
    \centering
    \small
    \caption{Instruction prompts and generated cases for generation modes of Levels 0, 1, 2, and 3.}
    \setlength{\tabcolsep}{3.5pt}
    \begin{tabular}{p{.16\textwidth} p{.38\textwidth} p{.41\textwidth}}
    \toprule
        \textbf{Generation Mode} &
        \textbf{Instruction Prompt} &
        \textbf{Generated News} 
        \\ \midrule
        
L0: No Generator Automation

(Human Written)
&
/
&
...The newly minted Duchess Of Sussex must never go to bed before the Queen on overnight trips, she must never give out an autograph and she must refrain from making personal comments or opinions on current events  - just to name a few, Now, a former Buckingham Palace staff has revealed another bizarre rule the duchess must follower ...
\\      
\midrule

L1: Human Assistance
&
\# Context

You are a journalist. Given a passage, please paraphrase it.

\# Specifications

The content should be the same. Do NOT generate any other word.

\# Given passage \textit{[Referred News]}
&
...The new Duchess of Sussex cannot retire for the night before the Queen during overnight stays, is prohibited from giving autographs, and must avoid sharing personal views on current affairs, among other rules. Recently, a former Buckingham Palace employee disclosed another peculiar rule that the duchess must adhere to ...
\\      
\midrule
L2: Partial Generator Automation
&
\# Context

You are a journalist. Given a passage, please rewrite it to make it more convincing.

\# Specifications

The content should be the same. The style should be serious, calm and informative. Do NOT generate any other word.

\# Given passage \textit{[Referred News]}
&
...As the newly appointed Duchess of Sussex, she is required to adhere to specific protocols, including the obligation to never retire for the night before the Queen during overnight trips, the prohibition on giving autographs, and the necessity to refrain from expressing personal comments or opinions on current events, among others. Recently, a former Buckingham Palace staff member disclosed another unusual rule that the Duchess must follow ...
\\     
\midrule
L3: Conditional Generator Automation
&
\# Context

You are a journalist. Given a title and some related sentences, please write a piece of news.

\# Specifications

The writing style should be serious and convincing. Do NOT generate any other word.

\# Title: \textit{[Referred News Title]}

\# Sentences: \textit{[News Key Sentences]}
&
...Among these are the expectations that she must never retire to bed before the Queen and is prohibited from giving autographs or expressing personal opinions on current events. While the specifics of the pasta ban remain unclear, it underscores the often stringent and peculiar customs that dictate life within royal circles. As the Duchess continues to adapt to her role, this latest decree adds to the unique challenges she faces ...
\\
    \bottomrule
    \end{tabular}
    \label{tab:generation_types_fig}
\end{table}

\begin{table}[h]
  \centering
  \small
  \caption{Statistics of the LLM-generated news items.}
    \begin{tabular}{cccccc}
    \toprule
    \multirow{2}[1]{*}{\textbf{Generator}} & \multicolumn{4}{c}{\textbf{Type of LLM-Generated News}} & \multirow{2}[1]{*}{\textbf{Total}} \\
\cmidrule{2-5}          & Real  & Fake L1 & Fake L2 & Fake L3 &  \\
    \midrule
    gpt-4o-mini & 14,248 & 4,592  & 4,592  & 4,592  & 28,024 \\
    Llama-3.1 & 14,248 & 4,592  & 4,592  & 4,592  & 28,024 \\
    \bottomrule
    \end{tabular}
  \label{tab:llm_data}
\end{table}

\subsection{Data Analysis \& Quality Check}
\label{sec:data analysis}

\begin{figure}[h]
	\centering
	\includegraphics[width=1\textwidth]{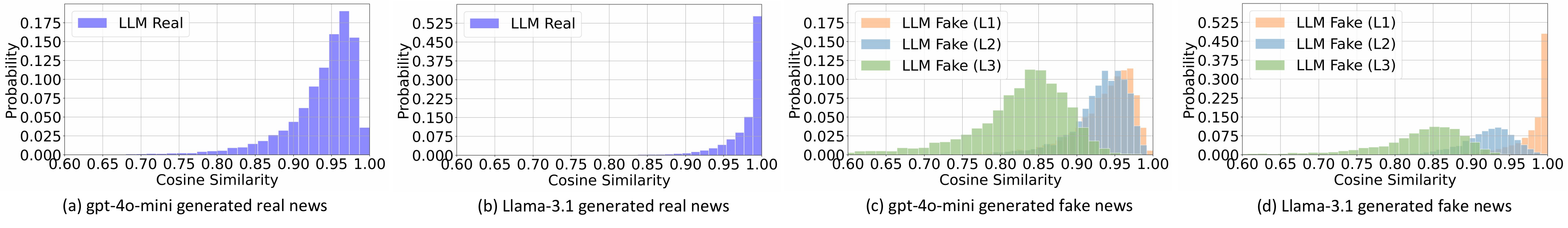}
	\caption{Cosine similarity distributions between embeddings of LLM-generated and human-written news.}
	\label{fig:llm_cos_sim}
\end{figure}

\noindent\textbf{Semantic Similarity Analysis.}
To analyze the semantic similarity between LLM-generated and human-written news, we utilized OpenAI's embedding model (\textit{text-embedding-3-small}) to extract semantic features for different types of news items. We then calculated the cosine similarity of the embeddings between each LLM-generated news item and its corresponding human-written counterpart.
From the statistical results in \figurename~\ref{fig:llm_cos_sim}, we observe that news generated using three different generation modes maintains a high level of semantic similarity with human-written news, indicating effective preservation of the original semantics. Specifically, as the level moves from L1 to L3, the semantic similarity gradually decreases. This trend aligns with our definition of generation mode levels: The higher the generation mode level, the stronger the LLM's involvement and freedom in generation.

\begin{figure}[h]
	\centering
	\includegraphics[width=0.6\textwidth]{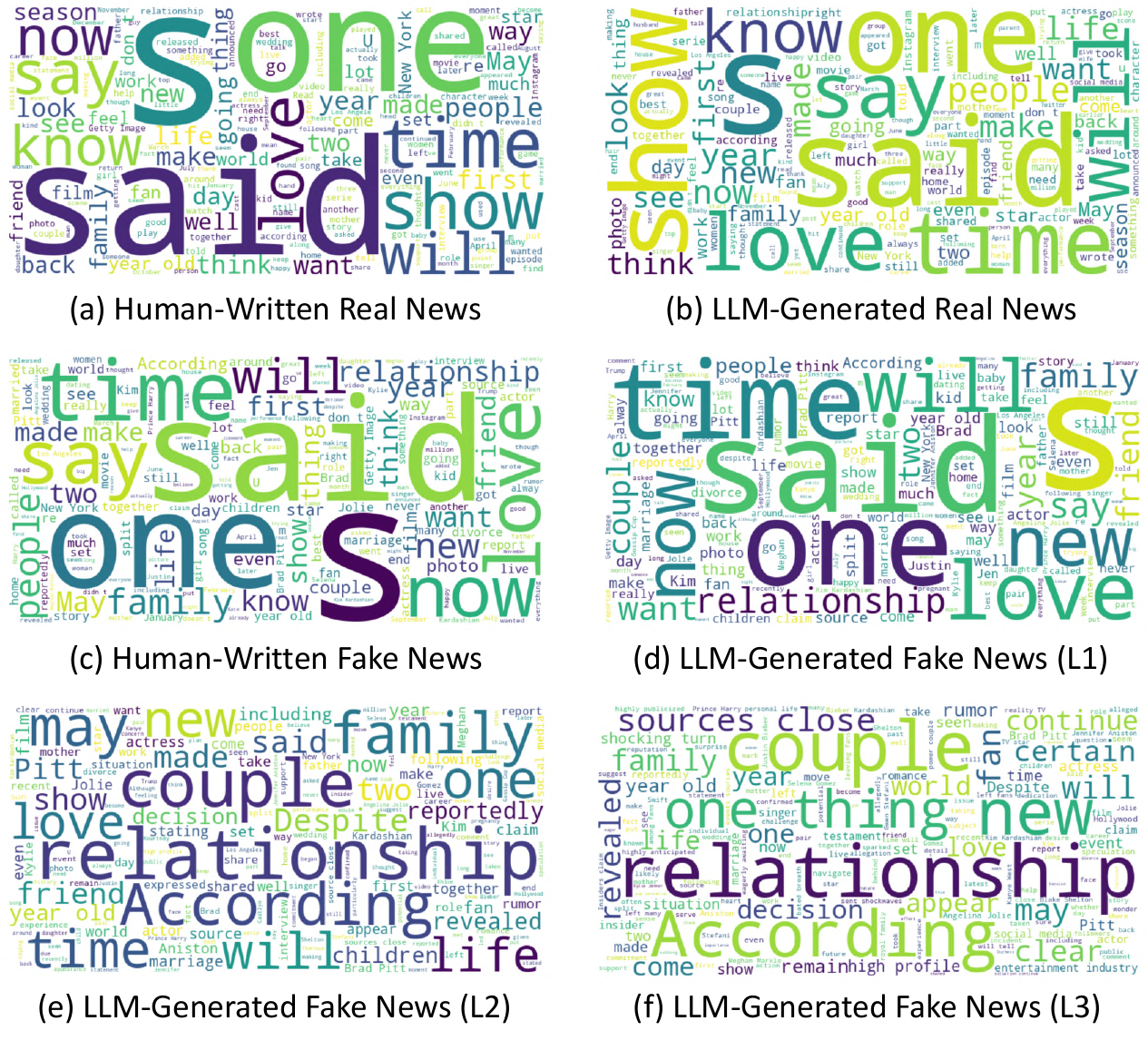}
	\caption{Wordclouds of human and generated news.}
	\label{fig:llama_wordcloud}
\end{figure}

\noindent\textbf{Stylistic Similarity Analysis.}
To analyze the stylistic similarities between LLM-generated and human-written news, we output word clouds for various types of news items. \figurename~\ref{fig:llama_wordcloud} presents the word clouds for human-written news and Llama-3.1 generated news. This figure shows distinct differences in the word clouds of different types of news, which highlights the stylistic diversity both between human-written and LLM-generated news and among the different generation modes of LLM-generated news. 

\noindent\textbf{Label Correctness Validation.}
To verify whether LLM-generated news items preserve the veracity labels of the referred human-written ones, we conduct a two-stage manual evaluation. Specifically, we conducted a stratified sampling on the constructed LLM-generated news dataset and obtained a subset of 400 news items (covering four different types of data generated by two distinct LLMs).
In the first stage, three human annotators independently evaluate event consistency between the LLM-generated news item and its corresponding human-written one. The final annotated labels are determined through a majority vote. As shown in \tablename~\ref{tab:content_consistency_fig}, the majority of LLM-generated news maintained consistency with their human-written counterparts, indicating the validity of the preserved veracity labels. In the second stage, for those cases where event descriptions were labeled as inconsistent, we conducted a manual review to verify the correctness of the given labels. Combining the results of the two stages, we conclude that the assigned labels of LLM-generated news guarantee sufficient validity.

\begin{table}[h]
  \centering
  \small
  \caption{Proportions of event-consistent LLM-human news item pairs in the human-annotated data.}
    \begin{tabular}{cccccc}
    \toprule
    \multirow{2}[1]{*}{\textbf{Generator}} & \multicolumn{4}{c}{\textbf{Type of LLM-Generated News}} \\
\cmidrule{2-5}          & Real  & Fake L1 & Fake L2 & Fake L3 \\
    \midrule
    gpt-4o-mini & 100\%  & 100\%  & 100\%  & 92\%  \\
    Llama-3.1 & 100\%  & 100\%  & 100\%  & 90\%  \\
    \bottomrule
    \end{tabular}
  \label{tab:content_consistency_fig}
\end{table}

\section{How Does Generated News Affect the Neural News Recommendation System?}
\label{sec:effects}

To investigate the impact of LLM-generated news on the news ecosystem, we conduct experiments at four distinct phases, as shown in \tablename~\ref{tab:exp_overview_fig}, based on the area intruded by LLM-generated news. Then we provide further analysis about the occurrence of truth decay from a familiarity perspective.

\begin{table}[h]
  \centering
  \small
  \caption{LLM-generated news's involvement in the experimental data in four phases. \ding{51} denotes that LLM-generated news is included and \ding{55} is the opposite. History: User-news interaction history. Candidates: News candidate list.}
    \begin{tabular}{ccccc}
    \toprule
    \multirow{2}[1]{*}{\textbf{Generated News Has Been In}} & \multicolumn{2}{c}{\textbf{Training Set}} & \multicolumn{2}{c}{\textbf{Test Set}} \\
\cmidrule{2-5} & History & Candidates & History & Candidates \\
    \midrule
    Nowhere (\S~\ref{sec:4.2}) & \ding{55}     & \ding{55}     & \ding{55}     & \ding{55} \\
    Candidates (\S~\ref{sec:4.3}) & \ding{55}     & \ding{55}     & \ding{55}     & \cellcolor{cyan!15} \cellcolor{cyan!15}\ding{51} \\
    History (\S~\ref{sec:4.4}) & \ding{55}     & \ding{55}     & \cellcolor{cyan!15}\ding{51}     & \cellcolor{cyan!15}\ding{51} \\
    Training Data (\S~\ref{sec:4.5}) & \cellcolor{cyan!15}\ding{51}     & \cellcolor{cyan!15}\ding{51}     & \cellcolor{cyan!15}\ding{51}     & \cellcolor{cyan!15}\ding{51} \\
    \bottomrule
    \end{tabular}
  \label{tab:exp_overview_fig}
\end{table}

\subsection{Experimental Setting}
\noindent\textbf{News Recommendation Models.}
To facilitate the evaluation and reflect the essential impact of LLM-generated news on neural news recommendation models, it is important to mitigate the influence of external factors beyond news content itself. Therefore, we carefully selected two typical and influential models in our experiments:
\textbf{1) LSTUR\citep{LSTUR}:} A neural news recommendation model that captures both long- and short-term user interests for improving recommendation accuracy. 
\textbf{2) NRMS\citep{NRMS}:} A neural news recommendation model that employs multi-head self-attention to encode news and user representations by modeling word interactions and user browsing behaviors.
Conventionally, we use a pre-trained language model, \textit{bert-base-uncased}~\cite{bert} for text representation.

\noindent\textbf{Evaluation Metrics.}
To assess the ranking of different types of news in the recommendation system, we used several evaluation metrics, including MRR, nDCG@K, and Ratio@K. MRR represents the mean reciprocal rank of the target type. nDCG@K measures the normalized discounted cumulative gain of the target type within the top-K results, while Ratio@K indicates the proportion of items in the target type within the top-K of the ranked results. 
During the evaluation process, to measure the ranking of a specific type, we set the labels of all data in the candidate list belonging to that type to 1, and those of other types to 0.
To facilitate the comparison between rankings of real and fake news, we introduce an indicator named \textbf{Relative Real Advantage (Relative RA/RRA)}:
\begin{equation}
    \text{RRA} = \frac{\text{Metric}_{\text{real}} - \text{Metric}_{\text{fake}}}{\text{Metric}_{\text{fake}}} \times 100\%,
\end{equation}
where Metric will be specified when RRA is calculated. RRA reflects the advantage of real news over fake news in terms of rankings. A positive value indicates that real news ranks higher than fake news, while a negative value suggests the opposite.

\noindent\textbf{Implementation Details.}
The PLM's parameters in news recommendation models are frozen during training, and the feature dimension extracted by the PLM is set to 768. The models' optimal learning rates are determined by grid search in [5e-3, 1e-6]. Each model is trained for 10 epochs using Adam optimizer~\citep{adam} with a batch size of 16, 
employing early stopping based on validation performance, where the best model in validation is selected for testing.
For NRMS, the self-attention network is configured with 8 heads, each having an output dimension of 8. For LSTUR, the CNN is set with 400 filters and a window size of 3. Both models utilize a dropout rate of 0.2. All experiments are repeated with five seeds.

\subsection{Phase 0: Generated News is Nowhere}
\label{sec:4.2}
To set a reference for subsequent LLM-involved experiments, we need to know the original neural news recommendation system purely trained on human-written news first. Specifically, we trained recommendation models using human-written news items and reported their ranking results of purely human-written candidate news items based on user history records of human-written news interactions.

\begin{table}[h]
  \centering
  \small
  \caption{Performance comparison of human-written fake and real news. RRA represents the relative advantage of real news over fake news. A positive RRA indicates that real news ranks higher, while a negative one suggests the opposite. HF/HR: Human-Written Fake/Real. Rec.: Recommender.}
    \begin{tabular}{ccccccc}
    \toprule
    \textbf{Rec.} & \textbf{Type} & \textbf{MRR}   & \textbf{nDCG@5} & \textbf{nDCG@10} & \textbf{Ratio@5} & \textbf{Ratio@10} \\
    \midrule
    \multirow{3}[1]{*}{LSTUR} & HF    & 17.15 & 45.63 & 43.91 & 43.37 & 42.11 \\
          & HR    & 18.83 & 54.37 & 56.09 & 56.62 & 57.89 \\
          & \textit{(RRA)} & \textit{9.80\%} & \textit{19.15\%} & \textit{27.74\%} & \textit{30.55\%} & \textit{37.47\%} \\
    \midrule
    \multirow{3}[1]{*}{NRMS} & HF    & 17.36 & 46.28 & 44.84 & 43.98 & 43.10 \\
          & HR    & 18.62 & 53.72 & 55.17 & 56.02 & 56.90 \\
          & \textit{(RRA)} & \textit{7.26\%} & \textit{16.08\%} & \textit{23.04\%} & \textit{27.38\%} & \textit{32.02\%} \\
    \bottomrule
    \end{tabular}
  \label{tab:stage_0_res}
\end{table}

From \tablename~\ref{tab:stage_0_res}, we observe that human-written real news consistently ranks higher than fake news. Across two recommendation models, real news outperforms fake news across five ranking metrics. This suggests that the news recommendation model tends to prioritize real news over fake news when only considering human data. This indicates that the original recommendation system, which only dealt with human-written news, somehow possesses a natural defense against fake news.

\subsection{Phase 1: Generated News Enters Candidates}
\label{sec:4.3}

Initially, LLM-generated news just enters the candidate news pool, waiting for users' interactions.
To simulate this stage where LLM-generated news is only in candidate lists, we trained recommendation models based on human-written news items and only included human-written news in the user history records, but the candidate list includes both human-written and LLM-generated news items.
For a candidate list first that includes $K$ human-written news items as in Phase 0, we add their corresponding LLM-generated news items into the list, resulting in a list of $2K$ items to guarantee a fair comparison between human and LLM news.
To evaluate the individual impact of different faking levels, we respectively prepare three types of lists, each of which exclusively contains LLM-generated fake news items of specified faking levels (i.e., L1, L2, or L3) plus human-written news items and LLM-generated real ones.
For example, if the human-written news is real, a generated real news item that has similar main content will be included; if it is fake and the generation mode is L1, a fake news item generated by referring to its content and the L1 prompt will be included.

\begin{table*}[h]
  \centering
  \small
  \caption{Ratio@5 comparison of different types of news. Relative RA (RRA) represents the relative advantage of real news over fake news. A positive Relative RA indicates that real news ranks higher, while a negative one suggests the opposite. ``vs. Baseline'' calculates the difference between the current Relative RA and the baseline Relative RA before LLM-generated news is introduced, for which \textcolor{blue}{a positive value} indicates an increased ranking advantage for real news, while \textcolor{red}{a negative one} suggests a decrease.
  HF/HR: Human-Written Fake/Real. GF/GR: LLM-Generated Fake/Real.
  }
  \setlength{\tabcolsep}{4.5pt}
    \begin{tabular}{ccccccccccccc}
    \toprule
    \multirow{2}[1]{*}{\textbf{Gen.}} & \multirow{2}[1]{*}{\textbf{Rec.}} & \multirow{2}[1]{*}{\textbf{Level}} & \multicolumn{4}{c}{\textbf{Types of A Joint View}} & \multicolumn{4}{c}{\textbf{Types of A Single View}} & \multicolumn{2}{c}{\textbf{Relative RA}} \\
\cmidrule(lr){4-7} \cmidrule(lr){8-11} \cmidrule(lr){12-13}  
&       &       & HF    & HR    & GF    & GR    & Fake  & Real  & Human & Generated & Value  & vs. Baseline \\
    \midrule
    \multirow{6}[1]{*}{\rotatebox{90}{Llama-3.1}} & \multirow{3}[1]{*}{LSTUR} & L1    & 23.07  & 25.63  & 23.77  & 27.54  & 46.84  & 53.17  & 48.70  & 51.31  & 13.51\% & \textcolor{red}{-17.04\%} \\
          &       & L2    & 23.21  & 24.19  & 26.99  & 25.61  & 50.20  & 49.80  & 47.40  & 52.60  & -0.80\% & \textcolor{red}{-31.35\%} \\
          &       & L3    & 23.74  & 25.84  & 23.06  & 27.37  & 46.80  & 53.21  & 49.58  & 50.43  & 13.70\% & \textcolor{red}{-16.85\%} \\
\cmidrule{2-13}          & \multirow{3}[1]{*}{NRMS} & L1    & 23.85  & 25.41  & 23.90  & 26.84  & 47.75  & 52.25  & 49.26  & 50.74  & 9.42\% & \textcolor{red}{-17.96\%} \\
          &       & L2    & 23.37  & 24.27  & 26.77  & 25.58  & 50.14  & 49.85  & 47.64  & 52.35  & -0.58\% & \textcolor{red}{-27.96\%} \\
          &       & L3    & 24.22  & 25.36  & 23.66  & 26.77  & 47.88  & 52.13  & 49.58  & 50.43  & 8.88\% & \textcolor{red}{-18.50\%} \\
    \midrule
    \multirow{6}[1]{*}{\rotatebox{90}{gpt-4o-mini}} & \multirow{3}[1]{*}{LSTUR} & L1    & 22.15  & 23.53  & 23.95  & 30.37  & 46.10  & 53.90  & 45.68  & 54.32  & 16.92\% & \textcolor{red}{-13.63\%} \\
          &       & L2    & 21.66  & 23.05  & 25.80  & 29.50  & 47.46  & 52.55  & 44.71  & 55.30  & 10.72\% & \textcolor{red}{-19.83\%} \\
          &       & L3    & 23.55  & 24.06  & 21.30  & 31.10  & 44.85  & 55.16  & 47.61  & 52.40  & 22.99\% & \textcolor{red}{-7.56\%} \\
\cmidrule{2-13}          & \multirow{3}[1]{*}{NRMS} & L1    & 23.37  & 23.91  & 23.45  & 29.27  & 46.82  & 53.18  & 47.28  & 52.72  & 13.58\% & \textcolor{red}{-13.80\%} \\
          &       & L2    & 22.39  & 23.00  & 26.65  & 27.96  & 49.04  & 50.96  & 45.39  & 54.61  & 3.92\% & \textcolor{red}{-23.46\%} \\
          &       & L3    & 24.24  & 25.24  & 19.95  & 30.57  & 44.19  & 55.81  & 49.48  & 50.52  & 26.30\% & \textcolor{red}{-1.08\%} \\
    \bottomrule
    \end{tabular}%
  \label{tab:stage_1_res}%
\end{table*}%

From \tablename~\ref{tab:stage_1_res}, we observe that:
\textbf{1)} The ranking advantage of real news over fake news decreases, and in some cases, fake news even surpasses real news. Comparing the RRA value with the baseline, we find that the introduction of LLM-generated fake news at all three faking levels reduces the ranking advantage of real news. 
At the faking level L2, when the generator is Llama-3.1, real news even ranks lower than fake news. This indicates that the introduction of LLM-generated news significantly suppresses the ranking of real news, which is termed Truth Decay below.
\textbf{2)} LLM-generated news consistently ranks higher than human-written news.
This observation aligns with the finding in \cite{dai2024neural}, highlighting the advantage of LLM-generated text over human-written text.

\subsection{Phase 2: Generated News Intrudes into User History}
\label{sec:4.4}
In Phase 2, LLM-generated news items have interaction records and thus generated news intrudes into user history.
As we did before, we first trained news recommendation models based on human-written news items and then used the mixed candidate lists in Phase 1 at the inference stage. 
However, the user history records now contain not only human-written but also LLM-generated news.
In this phase, we randomly replaced $t \times100$\% of human-written items in history records with their corresponding LLM-generated news items. $t \in[0, 1]$ is \textit{LLM ratio} which denotes intrusion degree of LLM-generated news.

\begin{figure}[h]
	\centering
	\includegraphics[width=1.0\textwidth]{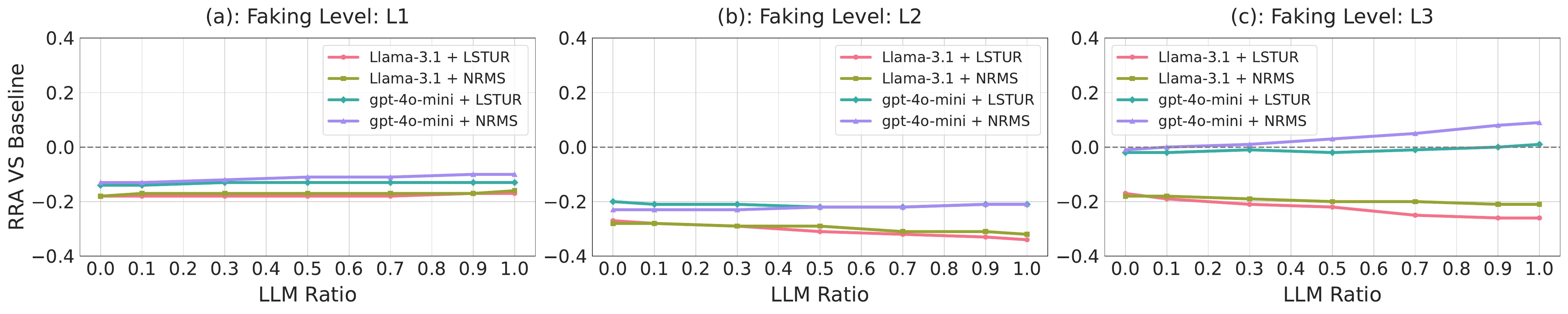}
	\caption{Variation in the difference between the current RRA and the baseline RRA as the LLM ratio in user history changes. A value below 0 indicates the RRA is lower than the baseline, while a value above 0 indicates the opposite. Due to space constraints, the RRA shown is calculated based on Ratio@5.}
	\label{fig:stage_2_res}
\end{figure}

From \figurename~\ref{fig:stage_2_res}, we observe that:
\textbf{1) }Generally, the RRA remains reduced compared to the baseline, indicating the truth decay phenomenon persists. This trend holds across varying LLM ratios in user history, except for the scenario where the faking level is L3 and the generator is gpt-4o-mini.
\textbf{2) }The overall trend of RRA varies across different faking levels. At L1, the RRA remains stable across different LLM ratios. At L2, when the generator is gpt-4o-mini, the RRA remains stable; In contrast, with Llama-3.1, the RRA decreases further as the LLM ratio increases. At L3, the RRA slightly increases with gpt-4o-mini, whereas with Llama-3.1, the RRA decreases further as the LLM ratio increases.

\subsection{Phase 3: Generated News Infiltrates Training Data}
\label{sec:4.5}
News recommendation models deployed in reality update the weights via routine retraining using data collected recently. This facilitates LLM-generated news to \textit{infiltrate} training data gradually and fundamentally affect model training.
For simulation, we constructed the training data using both human-written and LLM-generated news. Similar to what was done in Phase 2, we randomly replaced $t \times100$\% of human-written news items in training data with their corresponding LLM-generated ones. For inference, we employed mixed history records in Phase 2 with the same LLM ratio as in the training data, and mixed candidate lists used in both Phases 1 and 2.

\begin{figure}[h]
	\centering
    \includegraphics[width=1.0\textwidth]{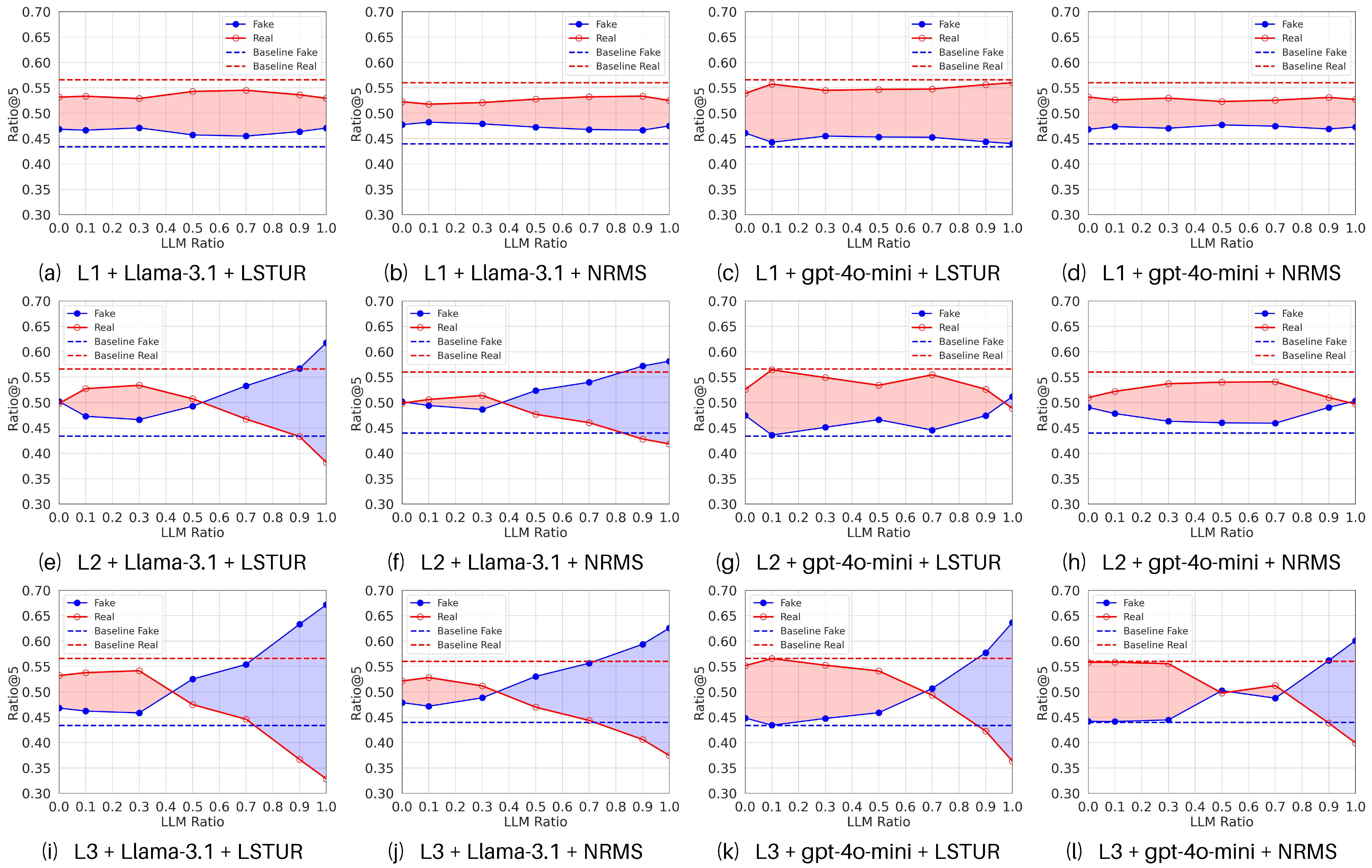}
	\caption{Variation in Ratio@5 for real and fake news. The \textcolor{red}{red dashed line} and \textcolor{blue}{blue dashed line} represent the Ratio@5 values for real and fake news, respectively, when no LLM-generated news is present. The \textcolor{red}{red shaded area} illustrates the extent to which real news leads fake news, while the \textcolor{blue}{blue shaded area} shows the extent to which fake news leads real news.}
	\label{fig:stage_3_res}
\end{figure}

\figurename~\ref{fig:stage_3_res} illustrates the changes in Ratio@5 for real and fake news as the LLM ratio varies. We observe that:
1) Overall, compared to the baseline, the presence of LLM-generated news in the training set consistently narrows the advantage of real news over fake news.
2) As the LLM ratio increases, this narrowing trend becomes more pronounced with higher faking levels. At L1, the advantage of real news remains stable. At L2, the real news advantage significantly diminishes and may even reverse. At L3, the ratio@5 of fake news eventually surpasses that of real news as the LLM ratio increases, indicating that the system favors fake news over real news.

\subsection{Analysis: Why Truth Decay Occurs}
We perform further analysis to provide a possible understanding of the underlying mechanisms by which truth decay occurs. As shown in Section~\ref{sec:data analysis}, there are no significant differences in overall semantics between LLM-generated and human-written news.
Therefore, other subtle differences noticed by recommendation models may contribute to such a ranking preference shift. 
Inspired by~\citep{dai2024neural,wangperplexity}, we focus on the backbone PLM used in news recommendation, as its intrinsic bias in language modeling would significantly affect content understanding and item scoring.
Specifically, we consider the perplexity metric because it indicates the model's familiarity with a text piece: The lower the perplexity, the more familiar the model is with the given text, possibly leading to better modeling. 

\begin{figure*}[h]
	\centering
	\includegraphics[width=1\textwidth]{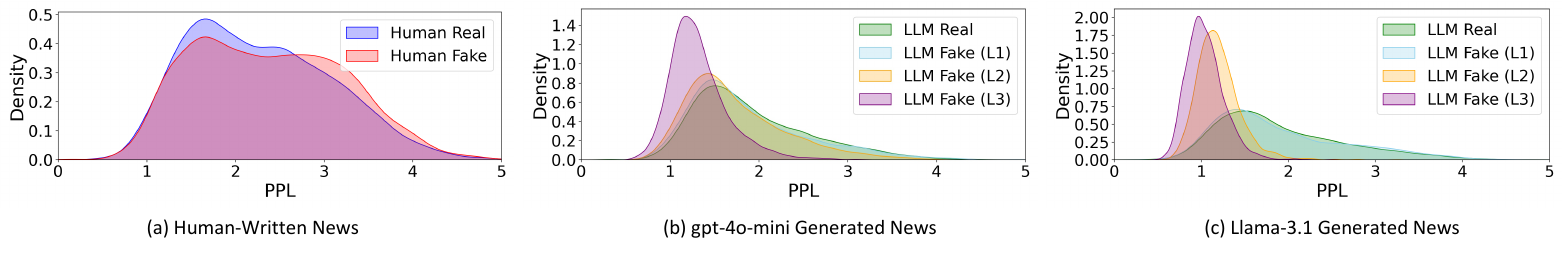}
	\caption{Perplexity distributions of LLM-generated and human-written news.}
	\label{fig:ppl_fig}
\end{figure*}

\figurename~\ref{fig:ppl_fig} illustrates the perplexities (PPLs) of LLM-generated and human-written news. We observe that:
1) In human-written news, real news exhibits lower perplexity than fake news, indicating an advantage of real over fake news, which is aligned with the real news advantages observed in Section~\ref{sec:4.2}.
2) Different from human-written news, LLM-generated fake news has lower perplexity than the real, corresponding to the emergence of truth decay when LLM-generated news is included. Moreover, the perplexity is lowered as the faking level increases, which provides an explanation about why L3 brings more significant truth decay than L1 and L2---with fewer constraints, generated news at L3 has lower perplexity and is more preferred by the backbone of news recommenders.

\section{Discussions}

\noindent\textbf{LLM-generated fake news exhibits a general threat to recommendation\-/based news ecosystem.} 
Our finding simulates the different extents and phases that LLM-generated news (both real and fake) is involved in news recommendation, and surprisingly show the ranking advantage of generated fake news. This suggests that as soon as our news ecosystem is penetrated by LLM-generated news, fake news creators could ride this wave and cause more significant harm than before. 
Current recommendation methods struggle to resist such pollution, and additional specific measures are needed.

\noindent\textbf{Possible countermeasures.}
Countermeasures to be taken require collaboration among the LLM providers, recommender systems, and news services. 1) For LLMs, apply robust safety alignment techniques~\cite{ganguli2022red,bai2022training} and mitigate the hallucination~\cite{tonmoy2024comprehensive} to avoid fake news production; 2) For recommender systems, evaluate and reduce the bias to LLM-generated text~\cite{zhou2024source} and factoring the credibility in news ranking~\cite{HDInt,bu2023combating}; And 3) for news services, apply LLM-generated text flagging~\cite{poger,liu2024preventing} and fake news detection~\cite{zoomout,arg}.

\noindent\textbf{Limitations.}
1) Though our analysis exposes the effects of LLM-generated fake news on news recommendation, how the effects ultimately influence human views remains unclear.
2) Our observations are based on a simulated recommendation environment, which excludes the factors from regulation measures~\cite{sharma2019combating,brown2021regulating}.
3) Our experiment is limited only to the observed news-faking approaches, which are (partially) based on human inputs. The impact of fully autonomous news faking (L5) remains unknown.
4) We do not test next-generation LLM-based recommendation models~\citep{liu2024rec,gao2024generative,li2023exploringfinetuningchatgptnews}.

\noindent\textbf{Ethical Considerations.}
We constructed a dataset via prompting LLMs to generate both real and fake news using publicly collected strategies.
To mitigate possible misuse risk, we used GossipCop as the referred human news data, which does not include recent news items, and will share the dataset for research needs only.

\section{Conclusion and Future Work}
We conducted an extensive simulated recommendation experiment and revealed the Truth Decay phenomenon in the recommendation-based news ecosystem, where real news is gradually losing its advantageous position in news ranking against fake news as LLM-generated news is involved.
This alarms us that the penetration and dominance of LLM-generated news may facilitate and escalate the fake news spread, raising a general threat to news ecosystems.
We advocate that new attention should be paid to this threat, and prevention is required to defend news integrity.

\noindent\textbf{Future Work.}
We plan to deepen our research by:
1) Simulating news faking in L4 and L5 modes to foresee its potential risk;
2) Extending the investigation to next-generation LLM-driven recommendation systems; and
3) Monitoring real-world LLM-based news faking campaigns to reveal the paradigm shift of news ecosystems.

\bibliography{main}

\begin{thebibliography}{67}
\providecommand{\natexlab}[1]{#1}

\bibitem[{An et~al.(2019)An, Wu, Wu, Zhang, Liu, and Xie}]{LSTUR}
Mingxiao An, Fangzhao Wu, Chuhan Wu, Kun Zhang, Zheng Liu, and Xing Xie. 2019.
\newblock \href {https://doi.org/10.18653/v1/P19-1033} {Neural News
  Recommendation with Long- and Short-term User Representations}.
\newblock In \emph{Proceedings of the 57th Annual Meeting of the Association
  for Computational Linguistics}, pages 336--345. Association for Computational
  Linguistics.

\bibitem[{Bai et~al.(2022)Bai, Jones, Ndousse, Askell, Chen, DasSarma, Drain,
  Fort, Ganguli, Henighan, Joseph, Kadavath, Kernion, Conerly, El-Showk,
  Elhage, Hatfield-Dodds, Hernandez, Hume, Johnston, Kravec, Lovitt, Nanda,
  Olsson, Amodei, Brown, Clark, McCandlish, Olah, Mann, and
  Kaplan}]{bai2022training}
Yuntao Bai, Andy Jones, Kamal Ndousse, Amanda Askell, Anna Chen, Nova DasSarma,
  Dawn Drain, Stanislav Fort, Deep Ganguli, Tom Henighan, Nicholas Joseph,
  Saurav Kadavath, Jackson Kernion, Tom Conerly, Sheer El-Showk, Nelson Elhage,
  Zac Hatfield-Dodds, Danny Hernandez, Tristan Hume, Scott Johnston, Shauna
  Kravec, Liane Lovitt, Neel Nanda, Catherine Olsson, Dario Amodei, Tom Brown,
  Jack Clark, Sam McCandlish, Chris Olah, Ben Mann, and Jared Kaplan. 2022.
\newblock \href {https://arxiv.org/abs/2204.05862} {Training a Helpful and
  Harmless Assistant with Reinforcement Learning from Human Feedback}.
\newblock \emph{Preprint}, arXiv:2204.05862.

\bibitem[{{BBC News}(2016)}]{google-fact-check}
{BBC News}. 2016.
\newblock Google News launches fact check label.
\newblock \url{https://www.bbc.com/news/technology-37657524}.
\newblock Accessed: 2025-04-12.

\bibitem[{Brown(2021)}]{brown2021regulating}
{\'E}tienne Brown. 2021.
\newblock Regulating the spread of online misinformation.
\newblock In \emph{The Routledge handbook of political epistemology}, pages
  214--225. Routledge.

\bibitem[{Brown et~al.(2020)Brown, Mann, Ryder, Subbiah, Kaplan, Dhariwal,
  Neelakantan, Shyam, Sastry, Askell, Agarwal, Herbert-Voss, Krueger, Henighan,
  Child, Ramesh, Ziegler, Wu, Winter, Hesse, Chen, Sigler, Litwin, Gray, Chess,
  Clark, Berner, McCandlish, Radford, Sutskever, and Amodei}]{gpt3}
Tom~B. Brown, Benjamin Mann, Nick Ryder, Melanie Subbiah, Jared Kaplan,
  Prafulla Dhariwal, Arvind Neelakantan, Pranav Shyam, Girish Sastry, Amanda
  Askell, Sandhini Agarwal, Ariel Herbert-Voss, Gretchen Krueger, Tom Henighan,
  Rewon Child, Aditya Ramesh, Daniel~M. Ziegler, Jeffrey Wu, Clemens Winter,
  Christopher Hesse, Mark Chen, Eric Sigler, Mateusz Litwin, Scott Gray,
  Benjamin Chess, Jack Clark, Christopher Berner, Sam McCandlish, Alec Radford,
  Ilya Sutskever, and Dario Amodei. 2020.
\newblock \href {https://arxiv.org/abs/2005.14165} {Language Models are
  Few-Shot Learners}.
\newblock \emph{Preprint}, arXiv:2005.14165.

\bibitem[{Bu et~al.(2023)Bu, Sheng, Cao, Qi, Wang, and Li}]{bu2023combating}
Yuyan Bu, Qiang Sheng, Juan Cao, Peng Qi, Danding Wang, and Jintao Li. 2023.
\newblock \href {https://doi.org/10.1145/3581783.3612426} {{Combating Online
  Misinformation Videos: Characterization, Detection, and Future Directions}}.
\newblock In \emph{Proceedings of the 31st ACM International Conference on
  Multimedia}, pages 8770--8780. Association for Computing Machinery.

\bibitem[{Caceres et~al.(2022)Caceres, Sosa, Lawrence, Sestacovschi,
  Tidd-Johnson, Rasool, Gadamidi, Ozair, Pandav, Cuevas-Lou, Parrish,
  Rodriguez, and Fernandez}]{caceres2022impact}
Maria Mercedes~Ferreira Caceres, Juan~Pablo Sosa, Jannel~A Lawrence, Cristina
  Sestacovschi, Atiyah Tidd-Johnson, Muhammad Haseeb~UI Rasool, Vinay~Kumar
  Gadamidi, Saleha Ozair, Krunal Pandav, Claudia Cuevas-Lou, Matthew Parrish,
  Ivan Rodriguez, and Javier~Perez Fernandez. 2022.
\newblock \href {https://doi.org/10.3934/publichealth.2022018} {The impact of
  misinformation on the COVID-19 pandemic}.
\newblock \emph{AIMS Public Health}, 9(2):262.

\bibitem[{Chen and Shu(2024{\natexlab{a}})}]{chen2024can}
Canyu Chen and Kai Shu. 2024{\natexlab{a}}.
\newblock Can {LLM}-Generated Misinformation Be Detected?
\newblock In \emph{The Twelfth International Conference on Learning
  Representations}.

\bibitem[{Chen and Shu(2024{\natexlab{b}})}]{chen2024combating}
Canyu Chen and Kai Shu. 2024{\natexlab{b}}.
\newblock \href {https://doi.org/10.1002/aaai.12188} {Combating Misinformation
  in the Age of LLMs: Opportunities and Challenges}.
\newblock \emph{AI Magazine}, 45(3):354--368.

\bibitem[{Chen et~al.(2024)Chen, He, Lin, Han, Wang, Cao, Sun, and
  Sun}]{spiral}
Xiaoyang Chen, Ben He, Hongyu Lin, Xianpei Han, Tianshu Wang, Boxi Cao, Le~Sun,
  and Yingfei Sun. 2024.
\newblock \href {https://doi.org/10.18653/v1/2024.acl-long.798} {Spiral of
  Silence: How is Large Language Model Killing Information Retrieval?{---}{A}
  Case Study on Open Domain Question Answering}.
\newblock In \emph{Proceedings of the 62nd Annual Meeting of the Association
  for Computational Linguistics (Volume 1: Long Papers)}, pages 14930--14951.
  Association for Computational Linguistics.

\bibitem[{Dai et~al.(2020)Dai, Sun, and Wang}]{FakeHealth}
Enyan Dai, Yiwei Sun, and Suhang Wang. 2020.
\newblock \href {https://doi.org/10.1609/icwsm.v14i1.7350} {{Ginger Cannot Cure
  Cancer: Battling Fake Health News with a Comprehensive Data Repository}}.
\newblock In \emph{Proceedings of the International AAAI Conference on Web and
  Social Media}, volume~14, pages 853--862. AAAI Press.

\bibitem[{Dai et~al.(2024{\natexlab{a}})Dai, Liu, Zhou, Pang, Ruan, Wang, Dong,
  Xu, and Wen}]{dai2024cocktail}
Sunhao Dai, Weihao Liu, Yuqi Zhou, Liang Pang, Rongju Ruan, Gang Wang, Zhenhua
  Dong, Jun Xu, and Ji-Rong Wen. 2024{\natexlab{a}}.
\newblock \href {https://doi.org/10.18653/v1/2024.findings-acl.421} {Cocktail:
  A Comprehensive Information Retrieval Benchmark with {LLM}-Generated
  Documents Integration}.
\newblock In \emph{Findings of the Association for Computational Linguistics:
  ACL 2024}, pages 7052--7074. Association for Computational Linguistics.

\bibitem[{Dai et~al.(2024{\natexlab{b}})Dai, Zhou, Pang, Liu, Hu, Liu, Zhang,
  Wang, and Xu}]{dai2024neural}
Sunhao Dai, Yuqi Zhou, Liang Pang, Weihao Liu, Xiaolin Hu, Yong Liu, Xiao
  Zhang, Gang Wang, and Jun Xu. 2024{\natexlab{b}}.
\newblock \href {https://doi.org/10.1145/3637528.3671882} {{Neural Retrievers
  are Biased Towards LLM-Generated Content}}.
\newblock In \emph{Proceedings of the 30th ACM SIGKDD Conference on Knowledge
  Discovery and Data Mining}, pages 526--537. Association for Computing
  Machinery.

\bibitem[{Devlin et~al.(2019)Devlin, Chang, Lee, and Toutanova}]{bert}
Jacob Devlin, Ming-Wei Chang, Kenton Lee, and Kristina Toutanova. 2019.
\newblock \href {https://doi.org/10.18653/v1/N19- 1423} {{BERT}: Pre-training
  of Deep Bidirectional Transformers for Language Understanding}.
\newblock In \emph{Proceedings of the 2019 Conference of the North {A}merican
  Chapter of the Association for Computational Linguistics: Human Language
  Technologies, Volume 1 (Long and Short Papers)}, pages 4171--4186.
  Association for Computational Linguistics.

\bibitem[{Fernandez and Bellog{\'\i}n(2020)}]{fernandez2020recommender}
Miriam Fernandez and Alejandro Bellog{\'\i}n. 2020.
\newblock \href {https://ceur-ws.org/Vol-2758/OHARS-paper3.pdf} {Recommender
  systems and misinformation: the problem or the solution?}
\newblock In \emph{Workshop on Online Misinformation- and Harm-Aware
  Recommender Systems}.

\bibitem[{Fern{\'a}ndez et~al.(2021)Fern{\'a}ndez, Bellog{\'\i}n, and
  Cantador}]{fernandez2021analysing}
Miriam Fern{\'a}ndez, Alejandro Bellog{\'\i}n, and Iv{\'a}n Cantador. 2021.
\newblock \href {https://arxiv.org/abs/2103.14748} {Analysing the Effect of
  Recommendation Algorithms on the Amplification of Misinformation}.
\newblock \emph{Preprint}, arXiv:2103.14748.

\bibitem[{Ganguli et~al.(2022)Ganguli, Lovitt, Kernion, Askell, Bai, Kadavath,
  Mann, Perez, Schiefer, Ndousse, Jones, Bowman, Chen, Conerly, DasSarma,
  Drain, Elhage, El-Showk, Fort, Hatfield-Dodds, Henighan, Hernandez, Hume,
  Jacobson, Johnston, Kravec, Olsson, Ringer, Tran-Johnson, Amodei, Brown,
  Joseph, McCandlish, Olah, Kaplan, and Clark}]{ganguli2022red}
Deep Ganguli, Liane Lovitt, Jackson Kernion, Amanda Askell, Yuntao Bai, Saurav
  Kadavath, Ben Mann, Ethan Perez, Nicholas Schiefer, Kamal Ndousse, Andy
  Jones, Sam Bowman, Anna Chen, Tom Conerly, Nova DasSarma, Dawn Drain, Nelson
  Elhage, Sheer El-Showk, Stanislav Fort, Zac Hatfield-Dodds, Tom Henighan,
  Danny Hernandez, Tristan Hume, Josh Jacobson, Scott Johnston, Shauna Kravec,
  Catherine Olsson, Sam Ringer, Eli Tran-Johnson, Dario Amodei, Tom Brown,
  Nicholas Joseph, Sam McCandlish, Chris Olah, Jared Kaplan, and Jack Clark.
  2022.
\newblock \href {https://arxiv.org/abs/2209.07858} {Red Teaming Language Models
  to Reduce Harms: Methods, Scaling Behaviors, and Lessons Learned}.
\newblock \emph{Preprint}, arXiv:2209.07858.

\bibitem[{Gao et~al.(2024)Gao, Fang, Tu, Yao, Chen, Ren, and
  Ren}]{gao2024generative}
Shen Gao, Jiabao Fang, Quan Tu, Zhitao Yao, Zhumin Chen, Pengjie Ren, and
  Zhaochun Ren. 2024.
\newblock \href {https://doi.org/10.1145/3589334.3645448} {Generative News
  Recommendation}.
\newblock In \emph{Proceedings of the ACM Web Conference 2024}, pages
  3444–--3453. Association for Computing Machinery.

\bibitem[{Giansiracusa(2021)}]{giansiracusa2021social}
Noah Giansiracusa. 2021.
\newblock \href {https://doi.org/10.1007/978-1-4842-7155-1_8} {\emph{Social
  Spread: Moderating Misinformation on Facebook and Twitter}}, pages 175--215.
\newblock Apress.

\bibitem[{Goldstein et~al.(2024)Goldstein, Chao, Grossman, Stamos, and
  Tomz}]{goldstein2024how}
Josh~A Goldstein, Jason Chao, Shelby Grossman, Alex Stamos, and Michael Tomz.
  2024.
\newblock \href {https://doi.org/10.1093/pnasnexus/pgae034} {How persuasive is
  AI-generated propaganda?}
\newblock \emph{PNAS Nexus}, 3(2):pgae034.

\bibitem[{Grinberg et~al.(2019)Grinberg, Joseph, Friedland, Swire-Thompson, and
  Lazer}]{science19}
Nir Grinberg, Kenneth Joseph, Lisa Friedland, Briony Swire-Thompson, and David
  Lazer. 2019.
\newblock \href {https://doi.org/10.1126/science.aau2706} {Fake news on Twitter
  during the 2016 US presidential election}.
\newblock \emph{Science}, 363(6425):374--378.

\bibitem[{Gulla et~al.(2017)Gulla, Zhang, Liu, {\"O}zg{\"o}bek, and
  Su}]{adressa}
Jon~Atle Gulla, Lemei Zhang, Peng Liu, {\"O}zlem {\"O}zg{\"o}bek, and Xiaomeng
  Su. 2017.
\newblock \href {https://doi.org/10.1145/3106426.3109436} {The Adressa dataset
  for news recommendation}.
\newblock In \emph{Proceedings of the International Conference on Web
  Intelligence}, pages 1042--1048. Association for Computational Linguistics.

\bibitem[{Hu et~al.(2024)Hu, Sheng, Cao, Shi, Li, Wang, and Qi}]{arg}
Beizhe Hu, Qiang Sheng, Juan Cao, Yuhui Shi, Yang Li, Danding Wang, and Peng
  Qi. 2024.
\newblock \href {https://doi.org/10.1609/aaai.v38i20.30214} {{Bad Actor, Good
  Advisor: Exploring the Role of Large Language Models in Fake News
  Detection}}.
\newblock In \emph{Proceedings of the AAAI Conference on Artificial
  Intelligence}, volume~38, pages 22105--22113. AAAI Press.

\bibitem[{Hu et~al.(2022)Hu, Wei, Zhao, and Wu}]{hu2022deep}
Linmei Hu, Siqi Wei, Ziwang Zhao, and Bin Wu. 2022.
\newblock \href {https://doi.org/10.1016/j.aiopen.2022.09.001} {Deep learning
  for fake news detection: A comprehensive survey}.
\newblock \emph{AI open}, 3:133--155.

\bibitem[{Huang et~al.(2023)Huang, Mckeown, Nakov, Choi, and
  Ji}]{huang2023faking}
Kung-Hsiang Huang, Kathleen Mckeown, Preslav Nakov, Yejin Choi, and Heng Ji.
  2023.
\newblock \href {https://doi.org/10.18653/v1/2023.acl-long.815} {Faking Fake
  News for Real Fake News Detection: Propaganda-Loaded Training Data
  Generation}.
\newblock In \emph{Proceedings of the 61st Annual Meeting of the Association
  for Computational Linguistics (Volume 1: Long Papers)}, pages 14571--14589.
  Association for Computational Linguistics.

\bibitem[{Jiang et~al.(2024)Jiang, Tan, Nirmal, and
  Liu}]{jiang2024disinformation}
Bohan Jiang, Zhen Tan, Ayushi Nirmal, and Huan Liu. 2024.
\newblock \href {https://doi.org/10.1137/1.9781611978032.50} {{Disinformation
  Detection: An Evolving Challenge in the Age of LLMs}}.
\newblock In \emph{Proceedings of the 2024 SIAM International Conference on
  Data Mining}, pages 427--435. SIAM.

\bibitem[{Jiang(2023)}]{TADI}
Junxiang Jiang. 2023.
\newblock \href {https://doi.org/10.18653/v1/2023.findings-emnlp.1047} {{TADI:
  Topic-aware Attention and Powerful Dual-encoder Interaction for Recall in
  News Recommendation}}.
\newblock In \emph{Findings of the Association for Computational Linguistics:
  EMNLP 2023}, pages 15647--15658. Association for Computational Linguistics.

\bibitem[{Konstan et~al.(1997)Konstan, Miller, Maltz, Herlocker, Gordon, and
  Riedl}]{cf1}
Joseph~A Konstan, Bradley~N Miller, David Maltz, Jonathan~L Herlocker, Lee~R
  Gordon, and John Riedl. 1997.
\newblock \href {https://doi.org/10.1145/245108.245126} {{GroupLens: applying
  collaborative filtering to Usenet news}}.
\newblock \emph{Communications of the ACM}, 40(3):77--87.

\bibitem[{Li et~al.(2022)Li, Zhu, Bi, Cai, Shang, Dong, Jiang, and Liu}]{MINER}
Jian Li, Jieming Zhu, Qiwei Bi, Guohao Cai, Lifeng Shang, Zhenhua Dong, Xin
  Jiang, and Qun Liu. 2022.
\newblock \href {https://doi.org/10.18653/v1/2022.findings-acl.29} {{MINER:
  Multi-interest matching network for news recommendation}}.
\newblock In \emph{Findings of the Association for Computational Linguistics:
  ACL 2022}, pages 343--352. Association for Computational Linguistics.

\bibitem[{Li et~al.(2023)Li, Zhang, and
  Malthouse}]{li2023exploringfinetuningchatgptnews}
Xinyi Li, Yongfeng Zhang, and Edward~C Malthouse. 2023.
\newblock \href {https://arxiv.org/abs/2311.05850} {{Exploring Fine-Tuning
  ChatGPT for News Recommendation}}.
\newblock \emph{Preprint}, arXiv:2311.05850.

\bibitem[{Li et~al.(2021)Li, Zhang, Du, Ma, and Wang}]{refutation}
Zongmin Li, Qi~Zhang, Xinyu Du, Yanfang Ma, and Shihang Wang. 2021.
\newblock \href {https://doi.org/10.1016/j.ipm.2020.102420} {Social media rumor
  refutation effectiveness: Evaluation, modelling and enhancement}.
\newblock \emph{Information Processing \& Management}, 58(1):102420.

\bibitem[{Liang et~al.(2024)Liang, Izzo, Zhang, Lepp, Cao, Zhao, Chen, Ye, Liu,
  Huang, Mcfarland, and Zou}]{monitoring}
Weixin Liang, Zachary Izzo, Yaohui Zhang, Haley Lepp, Hancheng Cao, Xuandong
  Zhao, Lingjiao Chen, Haotian Ye, Sheng Liu, Zhi Huang, Daniel Mcfarland, and
  James~Y. Zou. 2024.
\newblock \href
  {https://raw.githubusercontent.com/mlresearch/v235/main/assets/liang24b/liang24b.pdf}
  {Monitoring {AI}-Modified Content at Scale: A Case Study on the Impact of
  {C}hat{GPT} on {AI} Conference Peer Reviews}.
\newblock In \emph{Proceedings of the 41st International Conference on Machine
  Learning}, volume 235, pages 29575--29620. PMLR.

\bibitem[{Liu et~al.(2024{\natexlab{a}})Liu, Sheng, and Hu}]{liu2024preventing}
Aiwei Liu, Qiang Sheng, and Xuming Hu. 2024{\natexlab{a}}.
\newblock \href {https://doi.org/10.1145/3626772.3661377} {{Preventing and
  Detecting Misinformation Generated by Large Language Models}}.
\newblock In \emph{Proceedings of the 47th International ACM SIGIR Conference
  on Research and Development in Information Retrieval}, pages 3001--3004.
  Association for Computing Machinery.

\bibitem[{Liu et~al.(2024{\natexlab{b}})Liu, Yang, Du, Greene, Hurley, Lawlor,
  Dong, and Li}]{liu2024rec}
Dairui Liu, Boming Yang, Honghui Du, Derek Greene, Neil Hurley, Aonghus Lawlor,
  Ruihai Dong, and Irene Li. 2024{\natexlab{b}}.
\newblock \href {https://doi.org/10.1145/3627673.3679987} {{RecPrompt: A
  Self-tuning Prompting Framework for News Recommendation Using Large Language
  Models}}.
\newblock In \emph{Proceedings of the 33rd ACM International Conference on
  Information and Knowledge Management}, pages 3902–--3906. Association for
  Computing Machinery.

\bibitem[{Liu et~al.(2024{\natexlab{c}})Liu, Yang, Zhang, Kuang, Sun, Yang,
  Chen, Huang, and Wei}]{aipress}
Xiawei Liu, Shiyue Yang, Xinnong Zhang, Haoyu Kuang, Libo Sun, Yihang Yang,
  Siming Chen, Xuanjing Huang, and Zhongyu Wei. 2024{\natexlab{c}}.
\newblock \href {https://arxiv.org/abs/2410.07561} {AI-Press: A Multi-Agent
  News Generating and Feedback Simulation System Powered by Large Language
  Models}.
\newblock \emph{Preprint}, arXiv:2410.07561.

\bibitem[{Lo et~al.(2022)Lo, Dai, Xiong, Jiang, and Ku}]{victor}
Kuan-Chieh Lo, Shih-Chieh Dai, Aiping Xiong, Jing Jiang, and Lun-Wei Ku. 2022.
\newblock \href {https://doi.org/10.1145/3485447.3512246} {{VICTOR}: An
  Implicit Approach to Mitigate Misinformation via Continuous Verification
  Reading}.
\newblock In \emph{Proceedings of the ACM Web Conference 2022}, pages
  3511--3519. Association for Computing Machinery.

\bibitem[{Ma et~al.(2023)Ma, Liu, Xing, Qian, Lv, Yang, and Hu}]{PUNR}
Guangyuan Ma, Hongtao Liu, W~Xing, Wanhui Qian, Zhepeng Lv, Qing Yang, and
  Songlin Hu. 2023.
\newblock \href {https://doi.org/10.18653/v1/2023.findings-emnlp.559} {{PUNR:
  Pre-training with User Behavior Modeling for News Recommendation}}.
\newblock In \emph{Findings of the Association for Computational Linguistics:
  EMNLP 2023}, pages 8338--8347. Association for Computational Linguistics.

\bibitem[{P.~Kingma and Ba(2015)}]{adam}
Diederik P.~Kingma and Jimmy Ba. 2015.
\newblock \href {http://arxiv.org/abs/1412.6980} {{A}dam: {A} {M}ethod for
  {S}tochastic {O}ptimization}.
\newblock In \emph{International Conference on Learning Representations}.

\bibitem[{Park et~al.(2024)Park, Han, and Cha}]{park2024adversarial}
Sungwon Park, Sungwon Han, and Meeyoung Cha. 2024.
\newblock \href {https://arxiv.org/abs/2406.11260} {Adversarial Style
  Augmentation via Large Language Model for Robust Fake News Detection}.
\newblock \emph{Preprint}, arXiv:2406.11260.

\bibitem[{Pathak et~al.(2023)Pathak, Spezzano, and Pera}]{TWEB23Contribution}
Royal Pathak, Francesca Spezzano, and Maria~Soledad Pera. 2023.
\newblock \href {https://doi.org/10.1145/3616088} {Understanding the
  Contribution of Recommendation Algorithms on Misinformation Recommendation
  and Misinformation Dissemination on Social Networks}.
\newblock \emph{ACM Transactions on the Web}, 17.

\bibitem[{Sadeghi et~al.(2024)Sadeghi, Arvanitis, Padovese, Pozzi, Badilini,
  Vercellone, Wang, Huet, Fishman, Pfaller, Adams, and Wollen}]{newsguard}
McKenzie Sadeghi, Lorenzo Arvanitis, Virginia Padovese, Giulia Pozzi, Sara
  Badilini, Chiara Vercellone, Macrina Wang, Natalie Huet, Zack Fishman, Leonie
  Pfaller, Natalie Adams, and Miranda Wollen. 2024.
\newblock Tracking AI-enabled Misinformation.
\newblock
  \url{https://www.newsguardtech.com/special-reports/ai-tracking-center/}.
\newblock Accessed: 2024-12-06.

\bibitem[{SAE(2021)}]{sae}
SAE. 2021.
\newblock Taxonomy and Definitions for Terms Related to Driving Automation
  Systems for On-Road Motor Vehicles.
\newblock \url{https://www.sae.org/standards/content/j3016_202104/}.
\newblock Accessed: 2024-12-03.

\bibitem[{Saxena et~al.(2022)Saxena, Saxena, and Reddy}]{saxena2022fake}
Akrati Saxena, Pratishtha Saxena, and Harita Reddy. 2022.
\newblock \href {https://doi.org/10.1007/978-981-16-3398-0_16} {\emph{{Fake
  News Propagation and Mitigation Techniques: A Survey}}}, pages 355--386.
\newblock Springer.

\bibitem[{Sharma et~al.(2019)Sharma, Qian, Jiang, Ruchansky, Zhang, and
  Liu}]{sharma2019combating}
Karishma Sharma, Feng Qian, He~Jiang, Natali Ruchansky, Ming Zhang, and Yan
  Liu. 2019.
\newblock \href {https://doi.org/10.1145/3305260} {Combating Fake News: A
  Survey on Identification and Mitigation Techniques}.
\newblock \emph{ACM Transactions on Intelligent Systems and Technology}, 10.

\bibitem[{Sheng et~al.(2022)Sheng, Cao, Zhang, Li, Wang, and Zhu}]{zoomout}
Qiang Sheng, Juan Cao, Xueyao Zhang, Rundong Li, Danding Wang, and Yongchun
  Zhu. 2022.
\newblock \href {https://doi.org/10.18653/v1/2022.acl-long.311} {{Zoom Out and
  Observe: News Environment Perception for Fake News Detection}}.
\newblock In \emph{Proceedings of the 60th Annual Meeting of the Association
  for Computational Linguistics (Volume 1: Long Papers)}, pages 4543--4556.
  Association for Computational Linguistics.

\bibitem[{Shi et~al.(2024)Shi, Sheng, Cao, Mi, Hu, and Wang}]{poger}
Yuhui Shi, Qiang Sheng, Juan Cao, Hao Mi, Beizhe Hu, and Danding Wang. 2024.
\newblock \href {https://doi.org/10.24963/ijcai.2024/55} {{Ten Words Only Still
  Help: Improving Black-Box AI-Generated Text Detection via Proxy-Guided
  Efficient Re-Sampling}}.
\newblock In \emph{Proceedings of the Thirty-Third International Joint
  Conference on Artificial Intelligence}, pages 494--502. International Joint
  Conference on Artificial Intelligence Organization.

\bibitem[{Shu et~al.(2020)Shu, Mahudeswaran, Wang, Lee, and Liu}]{fakenewsnet}
Kai Shu, Deepak Mahudeswaran, Suhang Wang, Dongwon Lee, and Huan Liu. 2020.
\newblock \href {https://doi.org/10.1089/big.2020.0062} {{FakeNewsNet: A Data
  Repository with News Content, Social Context, and Spatiotemporal Information
  for Studying Fake News on Social Media}}.
\newblock \emph{Big data}, 8(3):171--188.

\bibitem[{Spitale et~al.(2023)Spitale, Biller-Andorno, and
  Germani}]{spitale2023ai}
Giovanni Spitale, Nikola Biller-Andorno, and Federico Germani. 2023.
\newblock \href {https://doi.org/10.1126/sciadv.adh1850} {AI model GPT-3 (dis)
  informs us better than humans}.
\newblock \emph{Science Advances}, 9(26):eadh1850.

\bibitem[{Su et~al.(2023)Su, Zhuo, Mansurov, Wang, and Nakov}]{su2023fake}
Jinyan Su, Terry~Yue Zhuo, Jonibek Mansurov, Di~Wang, and Preslav Nakov. 2023.
\newblock \href {https://arxiv.org/abs/2309.08674} {Fake News Detectors are
  Biased against Texts Generated by Large Language Models}.
\newblock \emph{Preprint}, arXiv:2309.08674.

\bibitem[{Sun et~al.(2024)Sun, He, Cui, Lei, and Lu}]{sun2024exploring}
Yanshen Sun, Jianfeng He, Limeng Cui, Shuo Lei, and Chang-Tien Lu. 2024.
\newblock \href {https://arxiv.org/abs/2403.18249} {Exploring the Deceptive
  Power of LLM-Generated Fake News: A Study of Real-World Detection
  Challenges}.
\newblock \emph{Preprint}, arXiv:2403.18249.

\bibitem[{Tonmoy et~al.(2024)Tonmoy, Zaman, Jain, Rani, Rawte, Chadha, and
  Das}]{tonmoy2024comprehensive}
S~M Towhidul~Islam Tonmoy, S~M~Mehedi Zaman, Vinija Jain, Anku Rani, Vipula
  Rawte, Aman Chadha, and Amitava Das. 2024.
\newblock \href {https://arxiv.org/abs/2401.01313} {A Comprehensive Survey of
  Hallucination Mitigation Techniques in Large Language Models}.
\newblock \emph{Preprint}, arXiv:2401.01313.

\bibitem[{Wang and Blei(2011)}]{cf2}
Chong Wang and David~M Blei. 2011.
\newblock \href {https://doi.org/10.1145/2020408.2020480} {Collaborative Topic
  Modeling for Recommending Scientific Articles}.
\newblock In \emph{Proceedings of the 17th ACM SIGKDD International Conference
  on Knowledge Discovery and Data Mining}, pages 448--456. Association for
  Computing Machinery.

\bibitem[{Wang et~al.(2025)Wang, Dai, Zhao, Pang, Zhang, Wang, Dong, Xu, and
  Wen}]{wangperplexity}
Haoyu Wang, Sunhao Dai, Haiyuan Zhao, Liang Pang, Xiao Zhang, Gang Wang,
  Zhenhua Dong, Jun Xu, and Ji-Rong Wen. 2025.
\newblock Perplexity Trap: PLM-Based Retrievers Overrate Low Perplexity
  Documents.
\newblock In \emph{The Thirteenth International Conference on Learning
  Representations}.

\bibitem[{Wang et~al.(2024{\natexlab{a}})Wang, Ma, Gao, Guo, Zhu, Fan, Lu, and
  Ng}]{wang2024mega}
Lionel~Z. Wang, Yiming Ma, Renfei Gao, Beichen Guo, Han Zhu, Wenqi Fan, Zexin
  Lu, and Ka~Chung Ng. 2024{\natexlab{a}}.
\newblock \href {https://arxiv.org/abs/2408.11871} {MegaFake: A Theory-Driven
  Dataset of Fake News Generated by Large Language Models}.
\newblock \emph{Preprint}, arXiv:2408.11871.

\bibitem[{Wang et~al.(2024{\natexlab{b}})Wang, Wang, Zhang, Wang, Liu, and
  Chen}]{HDInt}
Shoujin Wang, Wentao Wang, Xiuzhen Zhang, Yan Wang, Huan Liu, and Fang Chen.
  2024{\natexlab{b}}.
\newblock \href {https://doi.org/10.1145/3637528.3671944} {A Hierarchical and
  Disentangling Interest Learning Framework for Unbiased and True News
  Recommendation}.
\newblock In \emph{Proceedings of the 30th ACM SIGKDD Conference on Knowledge
  Discovery and Data Mining}, pages 3200--3211. Association for Computing
  Machinery.

\bibitem[{Wang et~al.(2022)Wang, Xu, Zhang, Wang, and Song}]{rec4mit}
Shoujin Wang, Xiaofei Xu, Xiuzhen Zhang, Yan Wang, and Wenzhuo Song. 2022.
\newblock \href {https://doi.org/10.1145/3485447.3512263} {{Veracity-aware and
  Event-driven Personalized News Recommendation for Fake News Mitigation}}.
\newblock In \emph{Proceedings of the ACM Web Conference 2022}, pages
  3673--3684. Association for Computing Machinery.

\bibitem[{Wang et~al.(2024{\natexlab{c}})Wang, Zhang, Koneru, Guo, Mingole,
  Sundar, Rajtmajer, and Yadav}]{wang2024reopening}
Xinyu Wang, Wenbo Zhang, Sai Koneru, Hangzhi Guo, Bonam Mingole, S~Shyam
  Sundar, Sarah Rajtmajer, and Amulya Yadav. 2024{\natexlab{c}}.
\newblock \href {https://arxiv.org/abs/2410.19250} {The Reopening of Pandora's
  Box: Analyzing the Role of LLMs in the Evolving Battle Against AI-Generated
  Fake News}.
\newblock \emph{Preprint}, arXiv:2410.19250.

\bibitem[{Wu et~al.(2019)Wu, Wu, Ge, Qi, Huang, and Xie}]{NRMS}
Chuhan Wu, Fangzhao Wu, Suyu Ge, Tao Qi, Yongfeng Huang, and Xing Xie. 2019.
\newblock \href {https://doi.org/10.18653/v1/D19-1671} {Neural News
  Recommendation with Multi-Head Self-Attention}.
\newblock In \emph{Proceedings of the 2019 Conference on Empirical Methods in
  Natural Language Processing and the 9th International Joint Conference on
  Natural Language Processing}, pages 6389--6394. Association for Computational
  Linguistics.

\bibitem[{Wu et~al.(2021)Wu, Wu, Qi, and Huang}]{PLM4REC}
Chuhan Wu, Fangzhao Wu, Tao Qi, and Yongfeng Huang. 2021.
\newblock \href {https://doi.org/10.1145/3404835.3463069} {Empowering News
  Recommendation with Pre-trained Language Models}.
\newblock In \emph{Proceedings of the 44th International ACM SIGIR Conference
  on Research and Development in Information Retrieval}, pages 1652--1656.
  Association for Computing Machinery.

\bibitem[{Wu et~al.(2020)Wu, Qiao, Chen, Wu, Qi, Lian, Liu, Xie, Gao, Wu, and
  Zhou}]{mind}
Fangzhao Wu, Ying Qiao, Jiun-Hung Chen, Chuhan Wu, Tao Qi, Jianxun Lian,
  Danyang Liu, Xing Xie, Jianfeng Gao, Winnie Wu, and Ming Zhou. 2020.
\newblock \href {https://doi.org/10.18653/v1/2020.acl-main.331} {{MIND: A
  Large-scale Dataset for News Recommendation}}.
\newblock In \emph{Proceedings of the 58th Annual Meeting of the Association
  for Computational Linguistics}, pages 3597--3606. Association for
  Computational Linguistics.

\bibitem[{Wu et~al.(2024)Wu, Guo, and Hooi}]{wu2024fake}
Jiaying Wu, Jiafeng Guo, and Bryan Hooi. 2024.
\newblock \href {https://doi.org/10.1145/3637528.3671977} {Fake News in Sheep's
  Clothing: Robust Fake News Detection Against LLM-Empowered Style Attacks}.
\newblock In \emph{Proceedings of the 30th ACM SIGKDD Conference on Knowledge
  Discovery and Data Mining}, pages 3367--3378. Association for Computing
  Machinery.

\bibitem[{Zellers et~al.(2019)Zellers, Holtzman, Rashkin, Bisk, Farhadi,
  Roesner, and Choi}]{zellers2019defending}
Rowan Zellers, Ari Holtzman, Hannah Rashkin, Yonatan Bisk, Ali Farhadi,
  Franziska Roesner, and Yejin Choi. 2019.
\newblock \href
  {https://proceedings.neurips.cc/paper_files/paper/2019/file/3e9f0fc9b2f89e043bc6233994dfcf76-Paper.pdf}
  {{Defending Against Neural Fake News}}.
\newblock In \emph{Advances in Neural Information Processing Systems},
  volume~32, pages 9054--9065. Curran Associates Inc.

\bibitem[{Zhang(2024)}]{wenzhou}
Gang Zhang. 2024.
\newblock Fabricated and altered the cause of cases: An MCN gang of online
  water army navy has been arrested.
\newblock
  \url{https://content-static.cctvnews.cctv.com/snow-book/index.html?item_id=1228812076456706991}.
\newblock Accessed: 2025-04-12.

\bibitem[{Zhao et~al.(2024)Zhao, Zhou, Li, Tang, Wang, Hou, Min, Zhang, Zhang,
  Dong, Du, Yang, Chen, Chen, Jiang, Ren, Li, Tang, Liu, Liu, Nie, and
  Wen}]{zhao2024survey}
Wayne~Xin Zhao, Kun Zhou, Junyi Li, Tianyi Tang, Xiaolei Wang, Yupeng Hou,
  Yingqian Min, Beichen Zhang, Junjie Zhang, Zican Dong, Yifan Du, Chen Yang,
  Yushuo Chen, Zhipeng Chen, Jinhao Jiang, Ruiyang Ren, Yifan Li, Xinyu Tang,
  Zikang Liu, Peiyu Liu, Jian-Yun Nie, and Ji-Rong Wen. 2024.
\newblock \href {https://arxiv.org/abs/2303.18223} {A Survey of Large Language
  Models}.
\newblock \emph{Preprint}, arXiv:2303.18223.

\bibitem[{Zhou et~al.(2020)Zhou, Mulay, Ferrara, and Zafarani}]{Recovery}
Xinyi Zhou, Apurva Mulay, Emilio Ferrara, and Reza Zafarani. 2020.
\newblock \href {https://doi.org/10.1145/3340531.3412880} {{ReCOVery: A
  Multimodal Repository for COVID-19 News Credibility Research}}.
\newblock In \emph{Proceedings of the 29th ACM International Conference on
  Information \& Knowledge Management}, pages 3205--3212. Association for
  Computing Machinery.

\bibitem[{Zhou et~al.(2024)Zhou, Dai, Pang, Wang, Dong, Xu, and
  Wen}]{zhou2024source}
Yuqi Zhou, Sunhao Dai, Liang Pang, Gang Wang, Zhenhua Dong, Jun Xu, and Ji-Rong
  Wen. 2024.
\newblock \href {https://arxiv.org/abs/2405.17998} {Source Echo Chamber:
  Exploring the Escalation of Source Bias in User, Data, and Recommender System
  Feedback Loop}.
\newblock \emph{Preprint}, arXiv:2405.17998.

\bibitem[{Zhu et~al.(2022)Zhu, Sheng, Cao, Nan, Shu, Wu, Wang, and
  Zhuang}]{m3fend}
Yongchun Zhu, Qiang Sheng, Juan Cao, Qiong Nan, Kai Shu, Minghui Wu, Jindong
  Wang, and Fuzhen Zhuang. 2022.
\newblock \href {https://doi.org/10.1109/TKDE.2022.3185151} {Memory-Guided
  Multi-View Multi-Domain Fake News Detection}.
\newblock \emph{IEEE Transactions on Knowledge and Data Engineering},
  35(7):7178--7191.

\end{thebibliography}

\appendix
\section{Additional Generation Modes}
\label{app:l4l5}

We attach the descriptions of L4/L5 to encourage future research.

\begin{itemize}
[nosep,leftmargin=1em,labelwidth=*,align=left,topsep=0pt]
    \item \textbf{L4: High Generator Automation.} This level allows the generator greater autonomy but in a controllable environment.
    It has full rights to decide the content, style, and theme of the generated news, but its external access is controlled by humans, and humans can stop the generation at any time.
    \item \textbf{L5: Full Generator Automation.} This level allows the generator to have full autonomy to create news content. Any decisions, including whether and when to generate news and element access and arrangements of content, style, and theme, are fully made by the generator itself.
\end{itemize}

\end{document}